\documentclass{ieeeaccess}
\usepackage{cite}
\usepackage{amsmath,amssymb,amsfonts}
\usepackage{algorithmic}
\usepackage{graphicx}
\usepackage{textcomp}

\usepackage[numbers]{natbib} 
\usepackage{makecell}
\usepackage{rotating}
\usepackage[table]{xcolor} 
\usepackage{glossaries}
\usepackage[export]{adjustbox}
\newacronym{nlp}{NLP}{natural language processing}
\newacronym{pos}{POS}{part-of-speech}
\newacronym{bow}{BOW}{bag-of-words}
\newacronym{oov}{OOV}{out-of-vocabulary}
\newacronym[plural=LMs,firstplural=language models (LM)]{lm}{LM}{language model}
\newacronym{tr}{TR}{text representation}
\newacronym[plural=FFNN,firstplural=Feed forward neural networks (FFNN)]{ffnn}{FFNN}{Feed forward neural network}
\newacronym[plural=CNN,firstplural=Convolutional neural networks (CNN)]{cnn}{CNN}{Convolutional neural network}
\newacronym[plural=RNN,firstplural=Recurrent neural networks (RNN)]{rnn}{RNN}{Recurrent neural network}
\newacronym{cbow}{CBOW}{continuous bag-of-words}
\newacronym{glove}{GloVe}{Global Vectors}
\newacronym{elmo}{ELMo}{Embeddings from Language Models}
\newacronym{lstm}{LSTM}{Long Short-Term Memory}
\newacronym{gru}{GRU}{Gated Recurrent Unit}
\newacronym{bilstm}{bi-LSTM}{bi-directional LSTM}
\newacronym{cove}{CoVe}{Context Vectors}
\newacronym{vdcnn}{VDCNN}{Very Deep Convolutional Neural Network}
\newacronym{gpt}{GPT}{Generative Pre-Training}
\newacronym{ulmfit}{ULMFiT}{Universal Language Model Fine-Tuning}
\newacronym{bert}{BERT}{Bidirectional Encoder Representations from Transformers}
\newacronym{roberta}{RoBERTa}{Robustly optimized BERT approach}
\newacronym{albert}{ALBERT}{A Lite BERT}
\newacronym{ernie}{ERNIE}{Enhanced Representation through kNowledge IntEgration}
\newacronym{electra}{ELECTRA}{Efﬁciently Learning an Encoder that Classiﬁes Token Replacements Accurately}
\newacronym{flops}{FLOPS}{Floating Point Operations Per Second}
\newacronym{mtdnn}{MT-DNN}{Multi-Task Deep Neural Network}
\newacronym{t5}{T5}{Text-To-Text Transfer Transformer}
\newacronym{xlm}{XLM}{Cross-lingual Language Models}
\newacronym{use}{USE}{Universal Sentence Encoder}
\newacronym{tnlg}{T-NLG}{Turing Natural Language Generation}
\newacronym{mtnlg}{MT-NLG}{Megatron-Turing Natural Language Generation}
\newacronym{deberta}{DeBERTa}{Decoding-enhanced BERT with disentangled attention}
\newacronym{luke}{LUKE}{Language Understanding with Knowledge-based Embeddings}
\newacronym{kepler}{KEPLER}{Knowledge Embedding and Pre-trained LanguagE Representation}
\newacronym{mass}{MASS}{MAsked Sequence to Sequence pre-training}
\newacronym{canine}{CANINE}{Character Architecture with No tokenization In Neural Encoders}
\newacronym{seq2seq}{Seq2Seq}{Sequence-to-Sequence}

\usepackage[hyphens]{url}

\def\BibTeX{{\rm B\kern-.05em{\sc i\kern-.025em b}\kern-.08em
    T\kern-.1667em\lower.7ex\hbox{E}\kern-.125emX}}
    
\begin{document}

\newcolumntype{?}[1]{!{\vrule width #1}}
\newcolumntype{P}[2]{%
  >{\begin{turn}{#1}\begin{minipage}{#2}\small\raggedright\hspace{0pt}}l%
  <{\end{minipage}\end{turn}}%
}

\history{Date of publication xxxx 00, 0000, date of current version xxxx 00, 0000.}
\doi{10.1109/ACCESS.2022.3205719}

\title{A Survey of Text Representation Methods and their Genealogy}
\author{\uppercase{Philipp Siebers}\authorrefmark{1},
\uppercase{Christian Janiesch\authorrefmark{2}, and Patrick Zschech}.\authorrefmark{3}}
\address[1]{Technische Universität Dresden, Helmholtzstr. 10, 01067 Dresden, Germany (e-mail: philipp.siebers@tu-dresden.de)}
\address[2]{TU Dortmund University, Otto-Hahn-Str. 12, 
44227 Dortmund, Germany (e-mail: christian.janiesch@tu-dortmund.de)}
\address[3]{Friedrich-Alexander-Universität Erlangen-Nürnberg, Schloßplatz 4, 91054 Erlangen, Germany (e-mail: patrick.zschech@fau.de)}

\markboth
{Siebers \headeretal: A Survey of Text Representation Methods and their Genealogy}
{Siebers \headeretal: A Survey of Text Representation Methods and their Genealogy}

\corresp{Corresponding author: Christian Janiesch (e-mail: christian.janiesch@tu-dortmund.de).}

\begin{abstract}
In recent years, with the advent of highly scalable artificial-neural-network-based text representation methods the field of natural language processing has seen unprecedented growth and sophistication. It has become possible to distill complex linguistic information of text into multidimensional dense numeric vectors with the use of the distributional hypothesis. As a consequence, text representation methods have been evolving at such a quick pace that the research community is struggling to retain knowledge of the methods and their interrelations. 
We contribute threefold to this lack of compilation, composition, and systematization by providing a survey of current approaches, by arranging them in a genealogy, and by conceptualizing a taxonomy of text representation methods to examine and explain the state-of-the-art. Our research is a valuable guide and reference for artificial intelligence researchers and practitioners interested in natural language processing applications such as recommender systems, chatbots, and sentiment analysis.
\end{abstract}


\titlepgskip=-15pt

\maketitle

\section{Introduction}
\label{sec:introduction}

Computational understanding of natural language is referred to as a particularly hard problem of science \cite{zhou_progress_2020} and sometimes it is even described as being ``simply too hard'' \cite{christiansen_language_2003}. Nonetheless, the research field of \textit{\gls*{nlp}} takes on this challenge to enable machines to fully understand human text and speech \cite{guo_review_2020}. The principal obstacle for a machine in \gls*{nlp} is the symbolic nature of text. Although a machine can process it, the meaning of language goes beyond what is represented \cite{goldberg_neural_2017}. To access the underlying information, traditional approaches compile texts according to grammatical rules or derived distance measures by comparing hand-crafted features and lexical information \cite{ferrone_symbolic_2020}. However, these methods are characterized by either poor scalability or their inability to capture more intricate linguistic features, that is semantic information. As a consequence, the research field arrived at a boundary and has grown stale in the decades following its conception as a testament to the complexity of the underlying problem \cite{9075398}.

Only recently,  the interest in the research field has been reignited by the explosion of available text data on the Internet, paving the way for novel data-driven approaches \cite{zhou_progress_2020, cambria_jumping_2014}. Artificial neural networks in particular enabled the distillation of linguistic information beyond the symbolic nature of text by representing words as multidimensional dense numeric vectors according to the distributional hypothesis, thereby encapsulating semantic meaning in a so-called \textit{\gls*{lm}}.

Owed to this achievement, \textit{\gls*{tr}} methods have been evolving at an unprecedented rate and the research community is struggling to keep up in providing an overview of the field. We contribute threefold to this lack of compilation, composition, and systematization.

\textbf{(1) Compilation.} While surveys of \gls*{tr} exist, they are characterized by a low coverage of existing methods, high-level explanations, and narrow perspectives (see also Section VI). We provide a comprehensive compilation of the current state-of-the-art of \gls*{tr} methods, detail their motivation, and describe how they implement the distributional hypothesis to create linguistically rich embeddings.

\textbf{(2) Composition.} Further, the genealogy of approaches that constitute the state-of-the-art is obfuscated by the rapid development of the field. We provide an annotated genealogy of \gls*{tr} methods. Specifically, we conduct an in-depth analysis from a conceptual and chronological viewpoint. We carve out evolutionary phases, streams, and differentiate the branches \textit{size}, \textit{context}, \textit{efficiency}, and \textit{multi-tasking}. Furthermore, we highlight methods that constitute historic \gls*{nlp} milestones due to their particular architectural design, eventuating in superior downstream task performance.

\textbf{(3) Systematization.} Lastly, as a means to provide conceptual guidance on how to judge and classify current and future \gls*{tr} methods, we extend beyond the schemata employed above and provide a conceptual taxonomy to classify \gls*{tr} methods along the dimensions \textit{architecture}, \textit{vocabulary}, \textit{representation}, \textit{domain dependency}, and \textit{training strategy}. We examine each dimension to systematize the current state-of-the-art.

In summary, our survey is the first to provide a comprehensive and detailed review of the state-of-the-art of \gls*{tr} and to propose a genealogy visualizing the interrelations and dependencies of the identified methods. Our research can be a valuable guide and reference for artificial intelligence researchers interested in \gls*{nlp} applications such as recommender systems, chatbots, and sentiment analysis.

Our research is structured as follows:
First, we elaborate the fundamentals of \gls*{nlp}, artificial neural networks, and \gls*{tr}. Next, we introduce our survey methodology based on the hermeneutic framework. Subsequently, we present a comprehensive list of \gls*{tr} methods and analyze their lineage to emphasize significant milestones. Lastly, we abstract from our survey and provide a conceptual taxonomy of \gls*{tr} methods, which we employ to examine the state-of-the-art. We close with an overview of related work as well as a conclusion and an outlook.

\section{Fundamentals}

\gls*{nlp} is a field of artificial intelligence that strives for a holistic computational understanding of natural language \cite{mote_natural_2012}. This is no trivial task as language is highly variable and ambiguous \cite{goldberg_neural_2017}. In fact, it is categorized as AI-complete, meaning that its resolution requires the ``synthesis of human-level intelligence'' with regard to natural language \cite{yampolskiy_turing_2013}.
Consequently, a machine must be able to understand each component of language, that is its phonology, morphology, syntax, semantics, and pragmatics. 
This is reflected in a set of specific challenges, called \textit{\gls*{nlp} tasks}, that provide a measure as to a machine's linguistic capabilities. These tasks can generally be solved by various methods. However, in recent years the research field has been dominated by machine learning approaches, in particular \textit{artificial neural networks} that enable deep learning \cite{9075398}. Importantly, such models cannot operate directly on discrete symbolic inputs. Hence,  \gls{tr} becomes a crucial second dimension of \gls*{nlp}.

\subsection{Natural Language Processing Tasks}

As previously outlined, computational language understanding can be roughly broken down into the \textit{lexical}, \textit{syntactic}, \textit{semantic}, and \textit{pragmatic} analysis of text through \gls*{nlp} tasks. Each linguistic layer therein unites previous layers and introduces a new level of complexity that entails more advanced methodologies. In general, the tasks in the lexical and syntactic layers are intermediate, that is they are not valuable on their own, but rather a means to an end for the resolution of more complex tasks \cite{manning_foundations_1999}. They deal with the accumulation of relevant insights on the inherent structural aspects of language. The tasks in the subsequent semantic and pragmatic layers can be considered higher level. They aim at understanding the meaning of language and the context it is used in. Note that while we present explicit \gls*{nlp} tasks in every linguistic layer in the following, state-of-the-art models aim to integrate the tasks of all layers implicitly.

\textbf{Lexical analysis.} On the lowest linguistic level, lexical analysis studies the structure of words. This encompasses their orthography, morphology, and, in case of spoken text, phonology. The fundamental task in this layer is the tokenization of text, that is its segmentation into smaller parts. These tokens typically represent words, characters, or concatenations thereof, called $n$-grams. The result of the tokenization is a vocabulary that constitutes the basis for virtually any other \gls*{nlp} task. Because of its central function, it is important to ensure that the tokenization process yields a high-quality vocabulary.
The segmentation of meaningful character $n$-grams (i.e., subwords) poses one challenge. An often-used algorithmic solution is byte-pair encoding, which greedily includes only the most frequent tokens in a corpus in the vocabulary \cite{radford_language_2019}. Higher-level representations, for example words or sentences, can then be constructed by combining the corresponding subwords.
Another challenge is the highly variant nature of text. It can be mitigated by a variety of preprocessing steps such as noise removal and text normalization. However, current approaches are becoming capable of modeling a large part of the intricacies of natural language in their billions of parameters and depend less and less on such modifications \cite{camacho-collados_word_2018}. 

\textbf{Syntactic analysis.} Syntactic analysis expands the linguistic scope from words to sentences to consider grammatical rules.
A central task in this layer is \textit{\gls*{pos}} tagging. It describes the traversal of a sentence to annotate the grammatical class of its constituent words. Typical \gls*{pos} are noun, verb, and adjective but can be more fine-grained \cite{gudivada_chapter_2018}.
The knowledge of the \gls*{pos} of a word has several possible applications, for instance, to improve word-sense disambiguation. Moreover, \gls*{pos} can be used in conjunction with data augmentation, for example to replace words with their synonym.
In addition, dependency parsing can reveal grammatical dependencies between words and phrases. To that end, a directed graph is created that holds dependency information, for example subject and object of a sentence.
A substantial challenge in this layer is the grammatical ambiguity of words \cite{kurdi_natural_2016}. Often, the same word assumes different \gls*{pos}. Another obstacle are multi-word expressions that do not obey standard grammatical rules \cite{goos_multiword_2002}. 

\textbf{Semantic analysis.} Semantic analysis aims at discerning the meaning of words and sentences in a language. The comprehension of a given sentence may require the understanding of preceding and succeeding sentences. Accordingly, the linguistic scope expands to involve relationships not only between words in a sentence, but between sentences themselves. A large variety of \gls*{nlp} tasks can be placed on the semantic layer, including text classification, word sense disambiguation, machine translation, or summarization. For a brief, non-exhaustive survey on this topic refer to \citet{9075398}.
The challenges for tasks in this layer can be grouped into two categories: semantic relationships between words and semantic relationships between words and the context they appear in. Instances of the former are synonyms and antonyms; instances of the latter are polysemy and multi-word expressions.

\textbf{Pragmatic analysis.} Lastly, pragmatic analysis aims at uncovering the meaning of language beyond what is expressed literally, distinguishing what is said from what is conveyed or accomplished \cite{korta_pragmatics_2020}. The linguistic scope broadens to world knowledge and common sense. 
Linguistic challenges in this layer are concepts such as hyponymy, hypernymy, and meronomy. Infusing a model with the ability to recognize these implications in the situations in which they are important to the meaning of a text is a hard task, not least because they commonly remain unspoken. 
\citet{yin_survey_2022} illustrate that a computational pragmatic understanding of natural language cannot yet be fully achieved and that merely atomistic solutions exist. Typically, they involve hand-crafted ontologies, for example WordNet\footnote{\url{https://wordnet.princeton.edu/}}, that maintain profound tree-like structures relating words through important concepts.

\subsection{Text Representation}

\gls*{tr} forms the basis of \gls*{nlp} \cite{wang_survey_2020}. It is concerned with the adequate encoding and formatting of natural language so that a machine can solve a \gls*{nlp} task. In order for a machine to derive meaning from text and solve more complex tasks, the unstructured, discrete symbols have to be transformed into a structured representation, i.e., numeric vectors. This process is called embedding. The two predominant embedding approaches are \textit{local} and \textit{distributional representations}. 

\textbf{Local representations.} Local representations are necessary for the initial conversion of symbols into vectors. Each vector dimension therein uniquely identifies a token of the vocabulary. However, for large corpora the proportion of unique tokens seen in a given text is usually much smaller than the amount of distinct tokens in the vocabulary, resulting in sparse vectors. This \textit{curse of dimensionality} leads to significant problems. Above all, it hinders a model from discovering relevant signals in the input data because it sees only a fraction of the enormous amount of possible feature combinations at any given time \cite{domingos_few_2012}. It is therefore essential to control the dimensionality of the vocabulary with growing corpora. 
Another implication of the alignment between the dimensions of the vectors with the vocabulary is the inability of local representations to express semantic or syntactic information. This is due to vector dimensions being orthogonal to each other and each individual token exhibiting the same distance to any other token in the vocabulary. As a positive effect, local representations are highly interpretable.

The most prevalent local representation method is \textit{one-hot encoding}. For a given text, each token is individually converted into a binary vector. The vectors are filled with zeroes except for a 1 at the dimension that corresponds to the token's index in the vocabulary. 
Analogue to symbolic representations, one-hot encodings do not provide information on the similarity of tokens. Therefore, they do not facilitate any kind of meaningful analysis of the underlying texts on their own. However, they enable vector and matrix calculations for discrete symbolic inputs, which constitutes the first step for distributional \gls*{tr} with artificial neural networks. This makes one-hot encodings essential for current \gls*{nlp} models.
In comparison, a \textit{count vector}, also known as \gls*{bow}, constitutes a local representation that is able to capture similarity information. It operates on a document level. To represent a document in a corpus, the vector representations are created as additive compositions of the one-hot encodings of their constituent tokens. Hence, if the same token appears more than once in a document, the summation of the corresponding one-hot encoding leads to document vectors that reflect frequency information.

\textbf{Distributional representations.} Distributional representations build on top of local embeddings to create vectors enriched with linguistic information and overcome the previously discussed disadvantages.
To address sparsity, local representations are projected into a shared multidimensional continuous vector space, thereby decoupling the vocabulary from the vector dimensionality to create dense representations. The dimensionality depends on the method that is used for the creation of the embeddings. For instance, if an artificial neural network is used, the dimensionality of the embeddings is prescribed by the hidden layer dimensionality of the network. In order to weave semantic and other linguistic information into the embeddings, tokens are projected according to their context. The contextualization is based on the distributional hypothesis \cite{harris_distributional_1954}, which states that the meaning of a word is defined by the distribution of its neighboring words, that is co-occurrence information. Concisely put by \citet{firth57synopsis}: ``you shall judge a word by the company it keeps''.
For example, the words ’apple’ and ’orange’ would be likely to appear in the same context and would thus occupy similar positions in the vector space. At the same time, words
that appear in different contexts would be moved further away from each other, for example ’apple’ and ’brick’. In conjunction, the vector space implicitly captures analogy relationships \cite{liu_visual_2018}. \citet{mikolov_efficient_2013} famously demonstrate that the vector operation \textit{vector(``king``) - vector(``man``) + vector(``woman``)} on trained embeddings results in a vector closest to \textit{vector(``queen``)}.
Eventually, the contextualization of the entire vocabulary according to the corpus produces distributional vector representations that abstract arbitrary and complex linguistic concepts across their dimensions. 
Unfortunately, this makes an interpretation intricate. More so, as each dimension is involved in multiple concepts at once \cite{liu_word_2020}.
\footnote{Distributional representation is not synonymous to distributed representation \cite{liu_word_2020}. Rather, it is a strict subset that focuses on contextual semantics \cite{ferrone_symbolic_2020}.}

\subsection{Artificial Neural Networks}

The application of artificial neural networks for \gls*{nlp} can be divided into the creation of distributional representations, that is the contextualization of textual units, and the resolution of downstream tasks, for example text classification. A wide variety of different models can be employed for either objective (for an introduction to machine learning and deep learning cf. also \cite{janiesch_em_2021}). Nonetheless, a smaller subset of algorithms dominates \gls*{tr} and \gls*{nlp} today, which we present in the following:

\textbf{\glspl*{ffnn}.} The \gls*{ffnn} is the simplest form of an artificial neural network \cite{10.1145/3374217}. The central element of an \gls*{ffnn} is the neuron, which controls the signal flow of the network. The neuron takes input signals, multiplies them with a weight, and adds a bias term. The output of the neuron is determined by passing the resulting value through an activation function. This enables the network to distinguish not linearly separable data \cite{10.1145/3374217}.
\gls*{ffnn} organize neurons into layers, in which the weights connect the output of all neurons in each layer to the input of the neurons in the following layer. Therefore, each layer  projects its input onto an $n$-dimensional vector space, where $n$ is the number of neurons in that layer. \gls*{ffnn} distinguish an input layer processing the original data representation, one or more hidden layers abstracting the output of the input layer by projecting it into a vector space corresponding to their dimensionality, and the output layer mapping the output of the previous layer to a number of neurons corresponding to the possible output values.

\textbf{\glspl*{cnn}.} \gls*{ffnn} are faced with a fundamental problem when being applied to discrete unstructured data: their fully-connected structure lets the number of parameters explode, impeding the ability of the network to learn relevant input signals \cite{bengio_neural_2003}. In response, a \gls*{cnn} connects the initial neurons only to parts of the input and models higher order dependencies in subsequent layers. Similar to \gls*{cnn} in computer vision that work on pixels as the atomic representation of images, \gls*{cnn} in the \gls*{nlp} domain use the building blocks of text, that is characters or words \cite{conneau_very_2017}. As \gls*{cnn} inherently create representations for the entire input sequence, their unmodified application for \gls*{nlp} is limited to tasks that require this coarse representation level. However, the limitation can be resolved by combining \gls*{cnn} with other model architectures \cite{peters_deep_2018}.

\textbf{\glspl*{rnn}.} \glspl*{rnn} are a broad family of artificial neural networks that specialize in the computation of sequential data \cite{young_recent_2018}. They are characterized by an explicit self-loop connection \cite{goodfellow_deep_2016}. This allows the model to deal with varying input lengths. The crucial difference to \gls*{ffnn} lies in the fact that the \gls*{rnn} shares its weight matrices. This enables it to generalize the detection of relevant signals in the input to any position and reduces parameter complexity \cite{goodfellow_deep_2016}.
The key element of any vanilla \gls*{rnn} is its hidden state. This is where the network implements the self-loop connection \cite{goodfellow_deep_2016}. The hidden state of the network thus can be compared to a memory that retains the most important information from the previous and current inputs in a compressed form, where the importance of information depends on the training task. Nonetheless, it is a lossy compression as an input vector of arbitrary length is reduced to a fixed length hidden vector \cite{goodfellow_deep_2016}.
Although \gls*{rnn} specialize in the computation of sequences, their architecture brings with it some issues. The unfolding of the parameter-shared \gls*{rnn} can lead to unstable gradients during the backwards pass with backpropagation through time and it may result in high computational resource requirements and memory usage.

\citet{hochreiter_long_1997} introduce a more sophisticated type of \gls*{rnn}, the \textit{\gls*{lstm}}, to address the problem of unstable gradients of the vanilla \gls*{rnn}. It introduces a dedicated memory mechanism with the cell state that is decoupled from the hidden state of the network and explicitly controlled by three gates: the forget gate, the input gate, and the output gate. This allows the memory to persist over time while being exposed to less encroaching operations than in the vanilla \gls*{rnn}. A variation of the \gls*{lstm} is the \textit{\gls*{gru}} \cite{cho_learning_2014}. It conceptually retains the gated architecture of the \gls*{lstm} but addresses the problem of high computational resource requirements by reducing tensor operations. Most notably, it drops the cell state of the \gls*{lstm} and returns to using the networks hidden state as its memory mechanism. Furthermore, the forget and input gate are combined into a single update gate. A second gate, the reset gate, provides the update gate with context for the generation of candidate values. It does so by merging a subset of features of the hidden state with the current input.

\textbf{Transformer.} The Transformer is a sequence-to-sequence model consisting of an encoder and a decoder block. The encoder block is made up of several layers with each layer containing multi-head attention, a parameter-shared \gls*{ffnn}, and normalization. The composition of the decoder block is similar, but inserts a masked multi-head attention layer to the front.
The principal aspect of the Transformer is that it relinquishes convolutions and recurrence and instead operates on attention. Attention can be described as a mechanism to highlight the importance of input features for a model in a given task \cite{galassi_attention_2021}. In the context of \gls*{tr} it can be thought of describing the linguistic composition of tokens in terms of syntactic and semantic similarities to all tokens in a sequence. This includes the influence of a token on itself.
As mentioned initially, the Transformer model augments the attention mechanism in two ways. On the one hand, the first layer in the decoder block employs masked attention, which prevents the calculation of attention scores for future tokens in a sequence. This is necessary as the decoder would otherwise be able to indirectly see the ground truth of the current token in a deep setting, preventing any learning effects \cite{devlin_bert_2019}. On the other hand, all attention scores are calculated with multiple different projections of the input to capture distinct input features. This is called multi-head attention.
The Transformer model constitutes a highly parallelizable architecture as it applies attention on the entire input sequence at all layers and shares the \gls*{ffnn} in each layer. Additionally, the Transformer excels in learning long-range dependencies in text by decreasing the path length between the positions of any input and output combination compared to other artificial neural network architectures \cite{NIPS2017_3f5ee243}. However, the attention mechanism is computationally expensive for long sequences \cite{cer-etal-2018-universal} and the encoder might be forced to learn irrelevant information of input sequences \cite{young_recent_2018}.

\section{Methodology of Literature Review}

We used the hermeneutic framework by \citet{boell_hermeneutic_2014} to guide our literature review. The hermeneutic framework construes a literature review as an iterative, interpretative process, in which literature retrieval and literature analysis alternate to facilitate the understanding of a research problem. Figure \ref{fig:hermeneutic framework} illustrates the elements that constitute the framework, that is the inner circle of \textit{search and acquisition} and the outer circle of \textit{analysis and interpretation}. 




\Figure[!t]()[width=0.6\textwidth, frame]{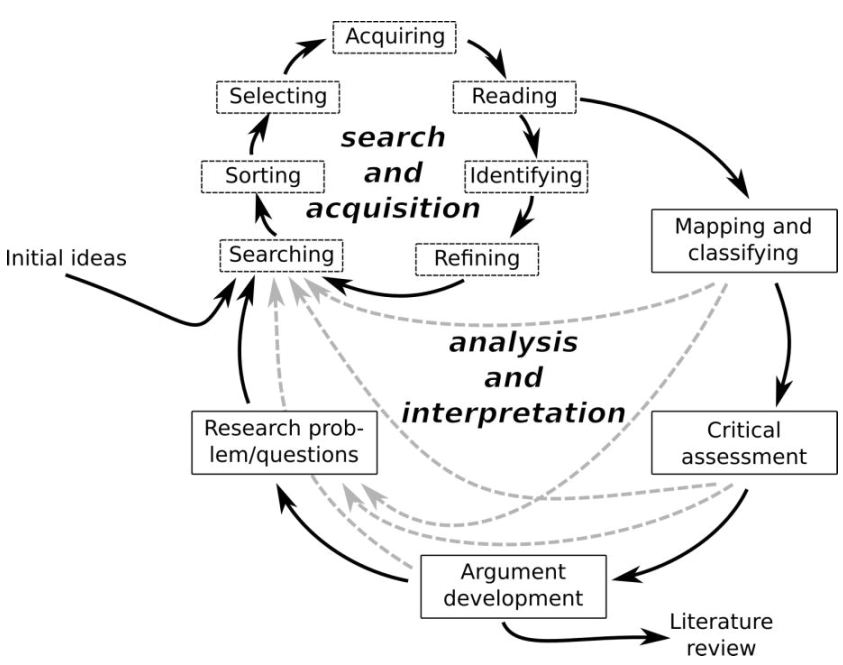}
   {Hermeneutic framework of \citet{boell_hermeneutic_2014}.\label{fig:hermeneutic framework}}


In the \textit{inner circle}, the body of literature is accumulated by identifying relevant publications and refining appropriate search terms. Further, the scholar advances their understanding of the problem domain by reassessing their interpretation of the problem's context and integrating new literature into it. In the \textit{outer circle}, the body of literature is analyzed on a broader scale. Publications are consolidated and compared among themselves with respect to content and methodology. Gaps in the current body of literature are identified to motivate further iterations of the hermeneutic circle. Gaps in the existing research are outlined to build an argument and highlight the research problem. There are several inter- and intra-circle linkages between the activities. They highlight how activities can influence each other while conducting the literature review.

We organized this process within a scope defined according to Cooper's taxonomy \cite{cooper_organizing_1988}: The \textit{focus} is placed on task-agnostic
\gls*{tr} method research outcomes of relevant papers, not their underlying theories nor their application or the methodology with which they were derived. The \textit{goal} of the review is the integration of the discovered information so that the reader gets an overview and understanding of recent approaches and their benefits and drawbacks. The \textit{organization} of the review is historical. The findings are presented in a manner that illustrates the evolution of \gls*{tr} along several dimensions. This way it becomes possible to retrace important conceptual changes and challenges, allowing for a comprehensive and descriptive explication of \gls*{tr} over time. The \textit{perspective} is neutral as information is presented objectively. The target \textit{audience} for this literature review are specialized scholars as prior knowledge in machine learning and data science is advisable. The \textit{coverage} cannot be exhaustive because of the hermeneutic character of the literature acquisition process. Regardless, a central/pivotal coverage better harmonizes with the historical organization of the review. Additionally, it allows for a more fine-grained implementation of the evolutionary aspect of this paper.

We considered a broad selection of scientific databases for our review, namely SpringerLink, arXiv, Web of Science, ACM Digital Library, IEEE Xplore, and ScienceDirect. We queried title, abstract, and keywords using the high-level term ``natural language processing'', its abbreviation ``NLP'', and the synonymous term ``computational linguistics'' as well as ``text mining'' to include all facets of \gls*{tr}. We excluded the term ``speech'' as speech recognition is a distinct field relying on a different set of methods. We conducted a backward search for each publication to identify additional relevant literature apart from the keyword search. As \citet{boell_hermeneutic_2014} point out, a literature search should incorporate sources apart from scientific databases. In response, we included the blogs \url{medium.com} and \url{towardsdatascience.com} in our literature search.

A \textit{saturation criterion} concludes each hermeneutic circle. Due to its subjectiveness, it is not defined by the framework but manifests as the absence of novel information over the previous iteration. Table \ref{tab:circles} summarizes the three search and acquisition circles we conducted.


\begin{table*}[ht]
\begin{tabular}{ |p{1.7cm}|p{12cm}| }
 \hline
 
 \rowcolor{darkgray} \multicolumn{1}{c}{\textcolor{white}{Scope}} & \textcolor{white}{Findings}
 \\
 \hline
First Circle: Overview of \acrshort*{nlp} & We established a general understanding of \gls*{nlp} before approaching the topic of \gls*{tr} in depth. Following \citet{boell_hermeneutic_2014}, we queried for surveys and reviews and found that the domain can be organized according to different \gls*{nlp} tasks, which in turn need effective methods (see respective section above). Furthermore, the literature depicted a gradual change from traditional learning to deep learning \gls*{nlp} models. \\ 
 \hline
 Second Circle: Overview of \acrshort*{tr} & We refined our focus to \gls*{tr}. Once more, we started with surveys and reviews. Novel key terms ``embedding'', ``distributional representation'', and ``distributed representation'' manifested. As we found ``computational linguistics'' and ``text mining'' to degrade the quality of search results for \gls*{tr}, we excluded those. In the course, different modalities of \gls*{tr} became apparent, of which local and distributional representations were found to be used exclusively in recent literature (see respective section above). Furthermore, we found \gls*{tr} to be shifting towards deep learning similarly to \gls*{nlp}, though at a quicker pace.  \\  
  \hline
 Third Circle: \acrshort*{tr} Methods & We transitioned to searching the blogs for only the most recent influential \gls*{tr} methods. After retrieving the corresponding publications, we conducted a recursive backwards search to retrace the history of \gls*{tr} methods and identify the seminal papers. Our search only uncovered one \gls*{tr} method that is not based on artificial neural networks. After this circle we observed signs of saturation. \\
\hline
\end{tabular}
\caption{Search and acquisition circles of the hermeneutic literature review.}
\label{tab:circles}
\end{table*}


The literature was retrieved in two steps as shown in Table \ref{tab:search_result}. After the initial search with the respective search terms, we analyzed title and abstract and found 167 
potentially relevant publications across all circles. After reading the full text, we discarded 83 
publications for a final set of 84 
publications.


\begin{table*}[h]
\begin{tabular}{|c?{1.3pt}c|c|c|c|}

\multicolumn{1}{c?{1.3pt}}{} & \cellcolor{darkgray} \textcolor{white}{1st Circle} & \cellcolor{darkgray} \textcolor{white}{2nd Circle} & \cellcolor{darkgray} \textcolor{white}{3rd Circle} & \cellcolor{darkgray} \textcolor{white}{$\sum$} \\ \Xhline{1.3pt}

\makecell{Identified Literature\\ \emph{(Title \& Abstract)}} & 70 & 41 & 64 & 175\\ \hline

\makecell{Retrieved Literature\\ \emph{(Full text)}} & 24 & 16 & 53 & 93 \\ \hline

\end{tabular}
\caption{Search result of the hermeneutic literature review.}
\label{tab:search_result}
\end{table*}


The initial literature search has been conducted in 2020 and was updated in 2022. The retrieved body of literature roughly covers a time span from 1995 to 2022. The distribution is heavily skewed towards recent publications across all hermeneutic circles. This underlines the radical change \gls*{tr} has undergone, rendering publications earlier than 2013 mostly irrelevant for the current progress in the field. This is due to two reasons: First, in that year a paradigm shift in the field of \gls*{tr} has led to the replacement of virtually all previous methods with superior artificial-neural-network-based approaches \cite{9075398}. Second, this paradigm shift has \mbox{(re-)}ignited interest in the research field, resulting in a substantial increase in publications. In consequence, we only include \gls*{tr} methods from 2013 or later in our review.

\section{Evolution of Text Representation Methods}

\subsection{Overview of Text Representation Methods}

In the following, we present and discuss the individual \gls*{tr} methods we identified. Our focus is placed on their motivation and how they implement the distributional hypothesis to create the embeddings. We organized the analysis based on the chronological development of the \gls*{tr} methods, however advancements on a particular method branch are grouped together. Note, that the date of publication is not always coherent with the date of the initial proposal of a \gls*{tr} method. In the following, we use the date of the initial proposal to arrange the \gls*{tr} methods.

\textbf{word2vec.}
\citet{mikolov_efficient_2013} propose two architectures, the \gls*{cbow} model and skip-gram model, commonly referred to as word2vec\footnote{word2vec is the name of the tensorflow implementation.}. The most important contribution of these models is the reduction of the computational complexity of calculating distributional representations. This allows the models to scale to large corpora and hence, to produce robust embeddings that accurately capture the linguistic relationships between words.

The \textit{\gls*{cbow}} model trains the embeddings by sliding a window over the text in the corpus and predicting each target word using its preceding and succeeding $n$ neighbours. To that end, a shared weight matrix is used to project the neighbouring words into the same vector space. Thereafter, the sum of the resulting vectors is used to predict the original target word. As a consequence, the vectors of the neighbouring words become aligned with the vector of the target word. However, during the projection of the words, any order information of the tokens is lost, hence the name of the model.

The \textit{skip-gram} model inverts the training process of the \gls*{cbow} model by using the projection of the target word to predict its $n$ preceding and succeeding neighbours. Because of the higher complexity of the skip-gram training task, \citet{10.5555/2999792.2999959} later introduce a number of improvements, most notably negative sampling. When predicting the neighbouring words on the basis of the target center word, this extension provides the model with additional randomly sampled negative words. The model is then trained to both minimize the distances from the target word vector to the vectors of the neighbouring words and maximize the distance to the negative samples.

\textbf{\Gls*{glove}.}
\citet{pennington_glove_2014} criticize that previous models like word2vec only account for local word co-occurrence information. With \gls*{glove}, they leverage the full statistical potential of corpora by calculating a context-aware global word-to-word co-occurrence matrix and transforming it into a distributed vector representation. Their approach is inspired by artificial-neural-network-based approaches, but it relies on simpler statistics. 
The co-occurrence matrix is populated at position $X_{ij}$, if the word $w_j$ appears in the context of $w_i$, with the context being a window of size $n$ around $w_i$. Subsequently, the embedding vectors are optimized according to a loss function in such a way, that the dot product of two arbitrary vectors approximate the log probability of the corresponding words in the co-occurrence matrix. To meet this requirement, the dimensions of \gls*{glove} embeddings need to capture meaningful information about the global context of a word and about the word itself.

\textbf{skip-thought.}
With skip-thought, \citet{kiros_skip-thought_2015} aim at creating downstream-task-independent and generic sentence representations. For that purpose, the skip-gram model is adapted to reconstruct the preceding and succeeding sentence to an input sentence. It uses a \gls*{gru} encoder to generate a target sentence representation from the input. During the training of the model, two \gls*{gru} decoders must predict each word of the preceding and succeeding sentence respectively. For each prediction, they can access the encoder representation of the sentence and the input token at the previous time-step. Finally, the sum of the log-probabilities of the two predicted sentences serves as the loss to condition the encoder representation. A limitation of the skip-thought model is its rather small vocabulary compared to other \gls*{tr} methods, in particular word2vec. To mitigate the problem of \textit{\gls*{oov}} words during inference, skip-thought maps each token of word2vec to the most similar token in the skip-thought vocabulary through linear regression. 

\textbf{Char-\gls*{cnn}.}
\citet{NIPS2015_250cf8b5} present char-\gls*{cnn}, a \gls*{cnn} that operates on character-level features. They argue that information can be derived from this raw input signal without the necessity for syntactic or semantic knowledge of the underlying language. Furthermore, they question the lack of task specificity of earlier \gls*{tr} methods. Char-\gls*{cnn} is trained on a supervised text classification task. \citet{NIPS2015_250cf8b5} use six convolutional layers. Furthermore, a variety of filter sizes is defined with the intent of capturing different $n$-gram compositions in the text.

\textbf{FastText.}
FastText \cite{bojanowski_enriching_2017} enriches word embeddings with subword information with the goal of leveraging morphological aspects of words without the need for morphological analysis. The model achieves this through the embedding of character $n$-grams complementary to the word embeddings. By doing so, non-trivial representations for \gls*{oov} words are possible by summing the corresponding $n$-gram vectors. Additionally, complex semantic relationships can be captured through the morphological similarities of words. This aids in the accurate representation of morphologically rich languages, for example the Finnish language. FastText builds on the word2vec model and uses either \gls*{cbow} or skip-gram with negative sampling. 
To enable a reliable representation of both rare and \gls*{oov} words, parameters responsible for the subword embedding are shared.
\citet{joulin_fasttextzip_2016} improve the FastText implementation by substantially reducing memory complexity. On the one hand, discriminative pruning retains only the best features, that is words and subwords, under the condition that the entire vocabulary remains covered. On the other hand, embeddings are compressed with a quantization algorithm and hashing is extended from subwords only -- as in the original implementation -- to subwords and words.

\textbf{char2vec.}
\citet{cao-rei-2016-joint} propose the char2vec model with the goal of applying unsupervised morphological analysis on character-level features. The approach is closely related to the FastText model as both models extend word2vec and incorporate the structural information of words. However, FastText does not explicitly aim to uncover morphological aspects of words. Further differences lie in the feature granularity on the one hand and the composition of word representations on the other hand. 
The learning task of char2vec is similar to the skip-gram model with negative sampling. However, it is adapted by a function that represents the target word as a composite of two half-words.
The function employs a \textit{\gls*{bilstm}} and a subsequent \gls*{ffnn} with an attention model. The two \gls*{lstm} read the characters of the target word from left to right and from right to left. Subsequently, the hidden states of both \gls*{lstm} are concatenated so that each concatenation represents a different split of the target word. With the goal of applying the embedding of the target word to the skip-gram training task, an \gls*{ffnn} is used to reduce the inflated vector dimensionality, resulting from the previous concatenation of two hidden states, back to its original size. An attention model then weighs the importance of each split of the word. Finally, the weighted vectors are summed and make up the input to the skip-gram task with negative sampling.

\textbf{context2vec.}
\citet{melamud_context2vec_2016} argue that fixed size context windows, which are used by previous \gls*{tr} methods, inhibit the representation of long range dependencies. Thus, the resulting embeddings reflect only a fraction of the information of a sentence. context2vec \cite{melamud_context2vec_2016} constitutes a model for the representation of contexts of variable length to solve this problem. Its architecture is based on the \gls*{cbow} model with negative sampling but replaces the averaging operation in the projection layer with a \gls*{bilstm} similar to the \gls*{bilstm} used in char2vec but operating on word-level. The resulting vector is a contextualized representation of the words surrounding the target word. It is used to optimize the \gls*{cbow} learning task with negative sampling.

\textbf{\gls*{vdcnn}.}
The \gls*{vdcnn} \cite{conneau_very_2017} highlights the benefit of deeper \gls*{cnn} architectures for \gls*{nlp}, similar to insights gained in the computer vision domain. The model pushes the number of convolutional layers from 6 (see char-\gls*{cnn}) to 29 while continually increasing performance on the jointly trained feature extraction and text classification task. The convolutional layers are organized in blocks as well as batch normalization and use max-pooling operations. Analogue to their computer vision counterparts VGG Net and ResNet, each pooling operation halves the resolution of the feature map and afterwards, the number of feature maps is doubled. The \gls*{vdcnn} uses a constant filter size of three throughout the network. These filters can be thought of to recognize character 3-grams in the first convolutional layer and learned compositions thereof in later convolutional layers. In a deep setting, this makes the approach more flexible than the char-\gls*{cnn}, which prescribes various filter lengths at the first layer.

\textbf{dict2vec.}
The premise of dict2vec \cite{tissier_dict2vec_2017} is that word embeddings can be improved upon using external resources, in particular natural language dictionaries. The definitions in the dictionary entries serve as additional curated contexts for all listed words and provide a mechanism to control the training of the network. Specifically, dict2vec uses the dictionary entries to generate so-called strong and weak word pairs. A strong word pair exists, if a word $w_1$ is in the definition of another word $w_2$ and $w_2$ is also in the definition of $w_1$. A weak pair exists, if $w_1$ is in the definition of $w_2$, but $w_2$ is not in the definition of $w_1$. 
At its base, dict2vec uses the skip-gram model, but extends it with two concepts, positive sampling and controlled negative sampling, that incorporate the discovered word-pair-relationships into the model training. Positive sampling selects a number of words that form weak and strong connections with the target word and introduces a loss so that the dot product of the vectors of the target word and its word pairs is minimized. Controlled negative sampling works similar to negative sampling described by \citet{10.5555/2999792.2999959}, but substitutes the error-prone random sampling of negative words with the sampling of words that do not form a weak or strong pair. This reduces the probability of taking coincidentally related words as negative examples.

\textbf{\Gls*{cove}.}
With \gls*{cove}, \citet{10.5555/3295222.3295377} draw inspiration from the success of transfer learning in the computer vision domain, where the generalization properties of \gls*{cnn} trained on a vast amount of data are transferred to other task-specific \gls*{cnn} to improve their downstream performance. The authors argue that the abundance of machine translation data can act as a catalyst for \gls*{nlp} models in a similar fashion. \gls*{cove} embeddings are created as a complement to previously created word embeddings, in particular \gls*{glove} embeddings, to provide them with a dynamic context representation. \gls*{cove} embeddings are trained on a supervised sequence-to-sequence machine translation task from English to German.
The model architecture consists of an encoder and a decoder. The encoder is a two-layer \gls*{bilstm} that uses word embeddings as its input and generates a sequence of hidden states at the final layer as its output. The decoder is a two-layer uni-directional \gls*{lstm} that attends over all encoder outputs.
In other terms, the final hidden states of the encoder reflect the bi-directional context of every word. Crucially, these representations need to generalize cross-lingual concepts, for example semantics, as they are indispensable for the decoder during the optimization of the machine translation task. Thus, it is the output of the encoder that constitutes the \gls*{cove} embeddings. 

\textbf{\Gls*{ulmfit}.}
\citet{howard_universal_2018} agree with \citet{10.5555/3295222.3295377} as to the importance of transfer learning for \gls*{nlp} and recognize more data as a driver of model performance. However, they propose language modeling\footnote{In a narrower sense, language modeling is concerned with the generation of the probability of a given sequence of words occurring in a sentence. To generate word probabilities, these \glspl{lm} are trained on large text corpora and thereby work as a tool to incorporate abundant information in a concise manner that is reusable in an out-of-sample context.} as the ideal training task instead of machine translation because it entails more data and captures many important linguistic facets. 
They propose \gls*{ulmfit} \cite{howard_universal_2018}, a method that enables efficient transfer learning for any \gls*{lm}. \gls*{ulmfit} consists of the three training phases (i.e., pre-training, fine-tuning, and classifier fine-tuning) and proposes the three transfer learning principles of discriminative fine-tuning, slanted triangular learning rates, and gradual unfreezing. 

In the first phase, an arbitrary \gls*{lm} is trained on the vast amount of data the domain offers.
In the second phase, the general text embeddings of the \gls*{lm} are adapted to the out-of-distribution downstream task data. During this process, discriminative fine-tuning prevents the model from overfitting to the smaller downstream task dataset by scaling down the learning rate for earlier layers. This preserves the general knowledge about language an \gls*{lm} captures at the first, less task-specific layers. At the same time, slanted triangular learning rates are used for an efficient adaptation of the model. This principle describes a quick initial increase and subsequent gradual decrease of the learning rate of the \gls*{lm}. It ensures that the \gls*{lm} first finds an adequate parameter space and then converges to the local optimum.
In the last phase, a classifier is fine-tuned. Training all classifier layers simultaneously introduces the risk of over- or underfitting the data. \gls*{ulmfit} uses gradual unfreezing to solve this issue. First, only the last, that is the least general layer, is trained and then, one by one, additional layers are unfrozen and trained to convergence. 

\textbf{\Gls*{use}.}
\citet{cer-etal-2018-universal} conjecture a higher transfer learning potential of sentence-level representations for downstream tasks in comparison to word embeddings. Moreover, they use unsupervised and supervised training to combine the benefits of more data and controlled linguistic training tasks respectively.
For the creation of general sentence representations, \citet{cer-etal-2018-universal} propose the \gls*{use}. Two encoder variants are considered, a Transformer encoder and a deep averaging network encoder. The former generates contextualized representations for every input token and composes a sentence representation afterwards, while the latter first composes the input embeddings to a sentence embedding and then contextualizes. The variants offer a trade-off between higher accuracy and lower computational complexity respectively. 

The encoders are trained by using multi-task learning on a skip-thought task, a supervised conversational input-response task, and a supervised classification task. The conversational input-response task has the model predict the response, given an input. During the classification task, the model is asked to assign a label to the input, that is ``entailment'', ``contradiction'', or ``neutral'', that describes the relationship between a hypothesis and a premise. According to these tasks, the model learns to embed various important aspects of language.
During training, the encoder is extended with deep artificial neural networks to form task-specific architectures. However, the parameters of the encoder are shared across all tasks to facilitate the generalization of its sentence embeddings.

\textbf{\Gls*{elmo}.}
\gls*{elmo} \cite{peters_deep_2018} mitigates the problem of polysemy for feature-based approaches by dynamically generating word embeddings for each input sequence. Furthermore, the method highlights how different layers capture specific linguistic aspects in a deep setting. \citet{peters_deep_2018} base their conceptual choices on the insights gained by previous \gls*{tr} methods. In particular, they train an \gls*{lm} and use a \gls*{cnn} as a low-level feature extractor.
\gls*{elmo} consists of three layers. The first layer is a character-level \gls*{cnn} that creates word embeddings of the input. These embeddings are contextualized by two layers, which each consist of a \gls*{bilstm} trained on a forward and a backwards language modeling task. In both layers, the hidden states of the two opposite \gls*{lstm} \glspl*{lm} are concatenated at each index of the input so that they are able to contextualize each word embedding with respect to the entire sequence.
The final \gls*{elmo} vectors are a learned, weighted sum of the output of the three layers of the network. The weighting scheme is downstream-task-specific because the first two layers of the model rather encode syntactic information, while the last layer encodes semantics.

\textbf{\Gls*{gpt}}
With \gls*{gpt}, \citet{radford_improving_2018} improve upon previous \gls*{lm}-based approaches by completely relying on attention operations \cite{NIPS2017_3f5ee243}. Specifically, the decoder part of the Transformer is used, which implies masked attention and hence an autoregressive \gls*{lm}. In this context, the direct interaction between a token and all previous tokens in a sequence is enabled, enhancing the model's capability to contextualize its embeddings \cite{zhou_progress_2020}. However, the focus of \gls*{gpt} not so much lies on fine-tuning this setup as on in-context learning and scalability. \citet{radford_improving_2018} train 12 decoder layers of the Transformer model on a series of downstream tasks with the help of task-specific input transformations and investigate the zero-shot performance of the \gls*{lm}, that is they apply it to a downstream task without any fine-tuning. 
With \textit{\gls*{gpt}-2}, \citet{radford_language_2019} further investigate the zero-shot capabilities of a pre-trained \gls*{lm} to determine, whether fine-tuning limits its expressiveness \cite{brown_language_2020}. To that end, a few changes are made to the original \gls*{gpt} architecture. Most notably, \gls*{gpt}-2 increases the number of layers to a total of 48, leading to a model size that is orders of magnitude larger than any previous \gls*{nlp} model, that is 1.5 billion parameters.
\textit{\gls*{gpt}-3} \cite{brown_language_2020} explores the scalability of the \gls*{gpt} model. The depth of a slightly tweaked version of \gls*{gpt}-2 is increased to 96 layers, resulting in 175 billion parameters. Furthermore, \citet{brown_language_2020} apply \gls*{gpt}-3 in a few-shot setting, that is conditioning the model to a downstream task by providing a small number of examples.

\textbf{\Gls*{bert}.}
Unlike \gls*{gpt}, \citet{devlin_bert_2019} only use the encoder block of the Transformer model \cite{NIPS2017_3f5ee243} to train a \gls*{bert} \gls*{lm}. Because the encoder uses self-attention, each token has direct access to all preceding and succeeding tokens. This allows \gls*{bert} to train a deeply bi-directional \gls*{lm}. 
However, the language modeling task of autoregressive \glspl*{lm} such as \gls*{elmo} or \gls*{gpt} cannot be used. Conditioning these \glspl*{lm} bi-directionally would enable a multi-layer network to trivially predict the target token \cite{devlin_bert_2019}. Therefore, \gls*{bert} trains an autoencoder \gls*{lm}, called masked \gls*{lm}, by corrupting the input and then trying to reconstruct it.
Precisely, 15\% of the tokens in the corpus are statically altered during the data pre-processing step. The alteration process either replaces a token with the $[MASK]$ token (80\%), a random token (10\%) or the original token (10\%). All tokens except for $[MASK]$ are then used to predict the original tokens.
In addition to the masked language modeling task, \gls*{bert} aims to incorporate knowledge about the relationships between sentences into the model. To that end, two sentences, separated by a special token $[SEP]$, form the input to the model. The model subsequently predicts, whether the second sentence follows the first one or is randomly drawn from the corpus. This task is called next sentence prediction. To facilitate a distinction between the tokens in the two segments, a segment encoding is added to the input embedding. With $[CLS]$, \gls*{bert} includes another special token. The token is placed at the beginning of every input sequence and is trained to attend to all tokens of the sequence. Hence, it is used to capture sequence-level information. 

\textbf{\Gls*{mtdnn}.}
\gls*{mtdnn} \cite{liu_multi-task_2019} combines language modeling with multi-task learning before fine-tuning. By adapting the pre-trained \gls*{bert} \gls*{lm} with several supervised \gls*{nlp} tasks, the model uses a large quantity of cross-task data without overfitting to a specific downstream task, resulting in regularization effects for more effective model fine-tuning. 
\gls*{mtdnn} adapts the \gls*{bert} \gls*{lm} by sequentially feeding input mini-batches of the different tasks to the network and updating the weights of all layers according to the loss function of the respective task. The model thus approximatively reduces the model error on all multi-task objectives simultaneously. \gls*{mtdnn} uses four task specific layers on top of the \gls*{lm}: single-sentence classification, pairwise text similarity, pairwise text classification, and relevance ranking. 
During the first task, a label corresponding to the sentiment of a sentence has to be predicted. The second task has the model predict a value that indicates the semantic similarity of two sentences. The third task is similar to the classification task employed in \gls*{use}. The last task is to rank the best answers to a given query.

\textbf{\Gls*{xlm}.}
\citet{NEURIPS2019_c04c19c2} criticize the monolingual focus of \gls*{tr} methods and the overrepresentation of the English language. They postulate that multilingual approaches can advance cross-lingual understanding and create high-quality text representations for low-resource languages and propose \gls*{xlm}.
The models align token representations across different languages by training on a shared, cross-lingual vocabulary. \gls*{xlm} implements three models each with a different training task: the autoregressive \gls*{lm} of \gls*{gpt}, the autoencoder masked \gls*{lm} of \gls*{bert}, and a translation \gls*{lm}. The translation \gls*{lm} transfers the masked \gls*{lm} to a supervised setting, in which two parallel sentences are concatenated for the input and aligned with a shared positional-embedding. In addition, a language embedding provides additional information for the model. Subsequently, input tokens are masked and predicted in accordance with the masked language modeling task.

\textbf{TransformerXL.}
\citet{dai_transformer-xl_2019} point out two key weaknesses of previous Transformer-based architectures: maximum context distance and context fragmentation. The former limitation restricts the modeling of relationships between tokens to the length of the input because contexts are no longer seen during training. The latter forces the model to train the first tokens in a given input sequence from scratch as context information from previous inputs cannot be accessed. Both limitations therefore result from a limited input sequence length. 
TransformerXL \cite{dai_transformer-xl_2019} trains an autoregressive \gls*{lm} that introduces the notion of recurrence to the Transformer architecture to overcome the above mentioned limitations. The outputs of the hidden layers of previous input sequences are cached and concatenated with the hidden states of the current input sequence to predict each token. Once the maximum memory capacity is reached, the oldest cached hidden states are discarded to free up space \cite{rae_compressive_2019}. 
In order to enable the recurrence mechanism, the positional encoding of the Transformer, which is statically implemented at the embedding layer, has to be made relative. Otherwise, the same indices would be used for different tokens when attending to previous sequences. \citet{dai_transformer-xl_2019} resolve this issue by extending the attention formula. Specifically, they add trainable biases that adjust for the distance between any two tokens.

\textbf{\Gls*{ernie}.}
\citet{sun_ernie_2019} argue that effective text representation methods should use more than just word co-occurrence information of the corpus. \gls*{ernie} adapts the masking strategy of the \gls*{bert} \gls*{lm} by extending it with phrase-masking and entity-masking. Phrases are defined as conceptual units consisting of characters or words. Entities are abstract or concrete concepts that can be denoted with a proper name. In this manner, the model incorporates prior knowledge about language and depends less on long context information.  
The \gls*{lm} is pre-trained with the help of a three-layer masking strategy. First, 15\% of the tokens in the input sequence are masked at random. In the second layer, entities of the input sequences are identified via lexical analysis and the corresponding tokens are masked. Lastly, phrases are identified and masked analogue to the entity masking process.
\citet{Sun_Wang_Li_Feng_Tian_Wu_Wang_2020} later introduced \textit{\gls*{ernie} 2.0}, a continuous multitask pre-training framework in line with the insights gained from their previous model proposition \gls*{ernie}. They extend the ways in which information other than co-occurence can be leveraged. \gls*{ernie} 2.0 enables the pre-training of an \gls*{lm} on different custom training tasks that can be extended by new tasks at any point in time.
It pre-trains a \gls*{bert} encoder on word-aware, structure-aware, and semantic-aware pre-training tasks that improve lexical, syntactic, and semantic capabilities respectively. 
Word-aware tasks consist of the adapted autoencoder \gls*{lm} introduced by \gls*{ernie}, the prediction of whether a token is capitalized and the prediction, whether a token appears at a different position of the same document.
Structure-aware tasks are the reordering of a permuted input sequence and a prediction, whether two sentences are adjacent, in the same document or from different documents.
Semantic-aware tasks include the prediction of the semantic or rhetorical relatedness of two text spans and a task similar to relevance ranking in the \gls*{mtdnn}.

\textbf{\Gls*{mass}.} \citet{pmlr-v97-song19d} use a full Transformer to condition a \gls*{gpt}-like decoder on a \gls*{bert}-like encoder. The encoder of their \gls*{mass} model is similar to \gls*{bert} as it masks part of the input. Nonetheless, the employed masking strategy differs. It is tailored to mask spans of text by replacing consecutive tokens with individual mask tokens $[M]$. 
The decoder of the \gls*{mass} model adopts the autoregressive \gls*{lm} of \gls*{gpt}. However, the unmasked tokens for the encoder are masked for the decoder to force the latter to rely more heavily on the input representations of the encoder and thus, the implicit linguistic knowledge encoded in them.

\textbf{XLNet.}
Despite their ability to model deep bi-directional contexts, autoencoder \gls*{lm} approaches face two major problems \cite{NEURIPS2019_dc6a7e65}: First, there is a discrepancy between pre-training and fine-tuning. Masked tokens are encountered during pre-training but never in a downstream task. Second, masked tokens are assumed to be independent of each other. They are predicted simultaneously without consideration for other masked tokens of the same sequence. 
XLNet \cite{NEURIPS2019_dc6a7e65} addresses these problems by creating an autoregressive \gls*{lm} that does not sacrifice bi-directionality, called permuted \gls*{lm}. This is achieved by maximizing the expected log-likelihood of all permutations of an input sequence. Naturally, the permutations include sequences, in which words that appeared after the target word in the original input, now appear before it and vice versa. Because the model parameters are shared the network hence learns bi-directional context information. 
XLNet permutes its factorization order, not the actual input to not deviate from inputs encountered during fine-tuning. This is achieved with attention masks. However, in a bi-directional context this introduces the problem of trivially predicting the target word \cite{NEURIPS2019_dc6a7e65}. To resolve this issue, XLNet introduces two streams of attention. The content stream, as used in the vanilla Transformer, is a contextualized representation of each input token. The query stream only contains the context representation for each input token. The query stream is used when predicting a token at the current position to hide the token identity and thus prevent a trivial prediction. The content stream is used in all other cases.
XLNet accepts the same input structure as \gls*{bert}, but the network is built on top of a TransformerXL backbone. Furthermore, the concept of relative embeddings introduced by TransformerXL is extended to the segment encoding of \gls*{bert}, mainly to improve generalization capabilities. 

\textbf{\Gls*{roberta}.}
\citet{liu_roberta_2019} argue that the original \gls*{bert} is severely under-trained and that the model can substantially improve in performance through effective and carefully crafted pre-training tasks.
To this end, \gls*{roberta} \cite{liu_roberta_2019} introduces small but effective changes to the vanilla \gls*{bert} model while leaving the core architecture almost identical. The first key change is making the masked language modeling task dynamic, that is generating the masking pattern each time an input sequence is passed into the network. This increases variance in the training data and thus helps the model to generalize better. The second key change is omitting the next sentence prediction task. This change effectively doubles the context length the model can represent because the input is no longer split into two potentially unrelated sequences. Furthermore, it prevents the introduction of unwanted noise during training \cite{spanbert}. \gls*{roberta} is also trained on more data with larger batches for a longer time and uses a larger byte-pair encoding vocabulary, that is smaller subword units than \gls*{bert}.

\textbf{SpanBERT.}
\citet{spanbert} accentuate the inability of many models to accurately represent multi-word expressions because they optimize the prediction of singular tokens during training.
The solution of SpanBERT \cite{spanbert} is straightforward. It adapts \gls*{bert} to mask spans of natural words in the input. The process is similar to phrase masking in \gls*{ernie}, that is each token of a span is replaced with an individual $[MASK]$ token. SpanBERT further introduces the span boundary task, in which the model has to recover a complete span using only its boundary tokens. This integrates span-level information into the respective boundary tokens. Lastly, like \gls*{roberta}, SpanBERT omits the next sentence prediction task.

\textbf{\Gls*{albert}}
The main contribution of \gls*{albert} \cite{lan_albert_2020} is its improved parameter efficiency over previous encoder-only Transformer models.
Through factorized embedding parametrization, \gls*{albert} disconnects the embedding matrix size from the hidden layer size of the model with a set of smaller matrices. This results in a parameter reduction when the hidden layer size is bigger than the embedding matrix size, allowing efficient scaling of the hidden matrix size according to the modeling needs. 
Additionally, \gls*{albert} shares all of its parameters across layers by default, further decreasing the amount of model parameters. Although this has a slight negative impact on downstream task performance, it stabilizes the network parameters during training \cite{lan_albert_2020}. 
Moreover, the next sentence prediction task of \gls*{bert} is not omitted but replaced with a sentence order prediction task. The authors argue that predicting whether a sentence belongs to the same document is too trivial of a task. In their proposed new task, the model has to ascertain whether the order of two consecutive sentences is correct or swapped.
Finally, \gls*{albert} implements the $n$-gram masking strategy of SpanBERT.

\textbf{Megatron-LM.}
When increasing the size of an \gls*{lm}, at some point memory limitations pose a substantial problem \cite{10.5555/3433701.3433727}. Therefore, \citet{shoeybi_megatron-lm_2020} explore the scalability of \glspl*{lm} by training Megatron-LM, which consists of multi-billion parameter versions of \gls*{bert} and \gls*{gpt}-2. In order to train these models, the authors rely on model parallelism techniques. 
Specifically, Megatron-LM distributes the calculation of general matrix multiplications across multiple workers. These matrix operations occur in several architectural components of the two previously mentioned models. 
In \gls*{ffnn}, the matrix multiplication of the weights with the input is split along the weight matrix columns to be distributed on several workers. During the output calculation, the respective weight matrix multiplication is parallelized along its rows.
Furthermore, Megatron-LM makes use of the parallel nature of the attention mechanism by splitting attention heads across several workers. However, the output of the attention heads has to be merged. For this, the respective matrix operation is distributed along the rows of the output. 
Lastly, Megatron-LM lets each worker optimize a set of parameters for layer normalization and dropout themselves to reduce communication requirements between workers. 

Building on the Megatron-LM backbone, \textit{\gls*{tnlg}} \cite{rosset_turing-nlg_2020} uses advances in distributed computing and distributed memory optimization technologies (i.e., DeepSpeed and Zer0) to train a Transformer-decoder-only model with 17 billion parameters.

\citet{smith_using_2022} go a step further and scale the same Megatron-LM variant to 530 billion parameters. Their model, the \textit{\gls*{mtnlg}}, is the largest non-sparse model at the time of writing. In order to be able to train a model of such size the authors pair the tensor-slicing of Megatron-LM with a complementary parallelization strategy across three dimensions. First, the model is divided into blocks which are distributed on parallel workers. Second, the individual model layers in each block are allocated to parallel workers. Third, the input data batches are distributed on parallel workers independently of the first two measures.
To efficiently coordinate the parallelisms of the model is to minimize the communication overhead between the parallel workers. To this effect, the topology of the hardware cluster is taken into account, that is  parallel workers requiring frequent communications are placed in the same or adjacent nodes.

\textbf{DistilBERT.}
Similarly to \gls*{albert}, DistilBERT \cite{sanh_distilbert_2020} focuses on reducing the computational complexity of a model while maintaining downstream task performance. 
It uses a setup called knowledge distillation, in which a small student model learns to replicate the behaviour of a larger teacher model. Both teacher and student are versions of \gls*{bert}, but the network depth of the student is halved. The student is initialized with the weights of the trained teacher and implements three loss functions to align their prediction behavior. A distillation loss aligns the probability distribution over all tokens. The masked language modeling task of \gls*{bert} is used in a supervised manner to align discrete token predictions, where the ground truth are the predictions of the teacher. Lastly, a cosine embedding loss aligns the embedding vectors for the tokens. 

\textbf{\Gls*{t5}}
\citet{JMLR:v21:20-074} agglomerate the most valuable insights gained by previous \gls*{tr} models and combine them in the \gls*{t5}, a large-scale text-to-text framework for the resolution of any \gls*{nlp} task.
\gls*{t5} relies on a full Transformer, as encoder- or decoder-only variants would limit downstream task applicability. To distinguish different tasks, the model is fed with a text string that indicates the task that has to be performed, for example ``TL;DR'' for summarization. The initial string is followed by prefixed input-strings in a task-specific structure. As implied by its name, the output of the model is always textual. 
The model also unifies the objective function to be a maximum likelihood for all tasks.
The encoder of \gls*{t5} uses a span masking task, in which a number of tokens are replaced by a singular mask token. 
At its base the decoder uses masked self-attention in an autoregressive manner to predict the spans. However, this setup can be limiting in conjunction with prefix information being passed to the model. For instance, in a machine translation task a sentence in the original language is provided so that the model knows what it has to translate. This information would be subject to masking as it forms part of the input. Thus, the model would only be able to access partial information for most time-steps to predict the translation. \gls*{t5} solves this issue by allowing the decoder to attend to any position of the prefixed input. Beyond that, the attention to previous positions remains prohibited.

\citet{DBLP:journals/corr/abs-2105-13626} adapt the \gls*{t5} model to operate on bytes instead of tokens, eliminating the need for vocabulary generation, text pre-processing, and tokenization. Their model \textit{ByT5} achieves competitive performance when masking longer spans of text and scaling the depth of the encoder.

\textbf{BART.} BART \cite{lewis_bart_2019} implements the standard Transformer-based neural machine translation architecture. However, the encoder fulfils the role of corrupting the input text according to an arbitrary function and the decoder learns to reconstruct the original in an autoregressive manner. Consequently, BART generalizes the masked \gls*{lm} of \gls*{bert} and the autoregressive \gls*{lm} of \gls*{gpt}. The approach allows for high flexibility in terms of pre-training input transformations. In particular, BART allows for changes to the length of the input text. This is used to mask spans of text similarly to Span\gls*{bert} but with the difference of inserting only a single $[MASK]$ token independently of the span length, hence incorporating knowledge about the amount of missing tokens on the decoder side. 

\textbf{Reformer.}
The Reformer \cite{kitaev_reformer_2020} joins the rank of \gls*{tr} methods that aim at reducing computational complexity such as DistilBERT. However, it is the first identified approach that targets the full Transformer architecture.
\citet{kitaev_reformer_2020} mainly carve out the attention calculation as a problematic factor in terms of computation. Their solution is the approximation of the key matrix of the attention formula. They reason that only the key values with the highest similarity to the query are required because the attention scores are subject to a softmax function. To sort the keys accordingly, locality-sensitive hashing is used. This hashing function maps each key vector to a bucket and ensures that similar vectors end up in the same bucket. Thus, the attention can be calculated efficiently for each respective bucket. The Reformer furthermore does not materialize the query matrix in memory for memory-efficiency at the cost of performance.

\textbf{Compressive Transformer.}
\citet{rae_compressive_2019} propose the Compressive Transformer. The model extends the TransformerXL with a new memory layer to model even longer context dependencies in text. 
In total, the Compressive Transformer accesses three layers of memory: the current sequence (vanilla Transformer), the TransformerXL memory, and the novel compressed memory. More specifically, instead of discarding old memories like the TransformerXL, a compression function is used to retain lossy representations. The compression function can be max/mean pooling, 1D convolution, dilated convolutions, or most-used, where memories are sorted by their average attention.
To facilitate the attention mechanism over previous sequences, the Compressive Transformer adopts the relative positional encoding of the TransformerXL.

\textbf{ProphetNet.} \citet{qi_prophetnet_2020} postulate that autoregressive LM overvalue local correlations of words at the cost of global dependencies. To address this issue, they adapt the autoregressive language modeling task to predict \textit{n} tokens for each token in a sequence. Hence, the model is forced to plan for future tokens. The change is implemented by extending the two-stream attention mechanism of XLNet to \textit{n}-stream self attention, where each stream \textit{i} is responsible for the prediction of the future \textit{i}-th token. ProphetNet further incorporates the masked \gls{lm} of \gls*{bert} on the encoder side of its \gls{seq2seq} architecture.

\textbf{\Gls*{electra}.}
\gls*{electra} \cite{clark_electra_2020} is similar to an inverse variant of DistilBERT, that is it uses a small network to optimize a large network. The focus of this approach lies on improving parameter efficiency as well. 
In particular, the masked \gls*{lm} of \gls*{bert} is extended with a replaced token prediction task. This novel task is implemented by concatenating two models, a generator and a discriminator. The generator is a reduced version of \gls*{bert} that trains on the masked language modeling task to generate and pass on plausible token replacements for 15\% of the input sequence. Subsequently, the discriminator, a larger \gls*{bert}, is initialized with the weights of the generator and has to predict for every token, whether it appears in the original sequence or has been altered by the generator. This procedure enables the discriminator to update the weights for all tokens per training step in comparison to only those of masked tokens in the masked language modeling task. Furthermore, no masked tokens are passed to the discriminator. This solves the pre-training/fine-tuning discrepancy of autoencoder \gls*{lm} discussed by XLNet.

\textbf{MPNet.} MPNet \cite{NEURIPS2020_c3a690be} leverages the advantages of permutative language modeling of XLNet and the masked \gls*{lm} of \gls*{bert} while mitigating their respective discrepancies between pre-training and fine-tuning. For this purpose, the authors create a masked and permuted language modeling task. In particular, MPNet can access 100\% of the positional information of a sequence in contrast to permutative language modeling and uses bidirectional context information between predicted tokens in contrast to the masked \gls*{lm}. This is implemented by permuting a token sequence and splitting it into three parts: non-masked tokens, mask-tokens, and the tokens that have been masked. Subsequently, self-attention is applied to the first two parts, while two-stream self-attention as in XLNet is applied to the entire token sequence.

\textbf{Funnel-Transformer.} The Funnel-Transformer \cite{NEURIPS2020_2cd2915e} improves computational efficiency by gradually reducing the hidden vector length with increasing model depth. The respective pooling function reduces the length of the hidden vector representation of the model with a sliding window mean pooling operation. Importantly, the resulting representation is not fed to the next layer directly but is only used to create the query vector for the self attention operation, while the full sequence representation is used for the key and value vectors. This makes the resulting pooled representation more context sensitive. However, due to the dimensionality reduction, the Funnel-Transformer loses the capability of representing individual input tokens. Since this might be required for the resolution of a downstream task, \citet{NEURIPS2020_2cd2915e} propose a decoder to recuperate expressive token-level representations. Concretely, the last hidden representation of the model is first upsampled to the original sequence length by repeating hidden vectors. Then, the last hidden vectors of the uncompressed part of the encoder are added to the upsampled hidden vectors. Finally, the result is passed through several additional Transformer layers to refine the token-level representations.

\textbf{BigBird.} \citet{NEURIPS2020_c8512d14} emphasize the low theoretical understanding of the self-attention operation of the original Transformer. Hence, they question the necessity of full self-attention, which scales quadratically with sequence length, for good \gls*{nlp} performance. 
BigBird uses sparse matrix calculations in three distinct attention patterns that retain the expressiveness and flexibility of the model while reducing computational complexity to be linear. 
The first pattern is random attention. Here, each query attends to a random number of randomly chosen keys. The goal is to approximate some characteristics of the full self-attention.
The second pattern is called window attention. It aims at capturing local relations between tokens as each token attends to a certain number of preceding and succeeding tokens.
The third pattern is global attention, in which a specific added or a chosen existing token attends to and is attended to by every other token in the input. This captures sequence-level information in the embedding of global tokens.

Roughly synchronous to BigBird, \citet{beltagy_longformer_2020} developed the similar \textit{Longformer} model, which exploits the lower computational complexity of the sparse attention operation to drastically increase the processable input sequence length. It uses the same three sparse attention patterns as BigBird, with the exception that dilated windows are used to further increase the receptive field of the model with increasing model depth at no additional computational cost.

\textbf{\Gls*{deberta}.} \gls*{deberta} \cite{he2021deberta} refines the token embedding strategy of \gls*{bert} in two ways. First, each token is explicitly represented with $n$ relative position embeddings to all other tokens in a sequence in addition to a content embedding. The positional and content vectors are summarized in matrices and used to calculate three attention scores of a token towards another, that is  content-to-content, content-to-position, and position-to-content. Content- and position-specific query and key projection matrices are used respectively. The above strategy captures crucial linguistic information on the relatedness and importance of tokens in a sequence. Second, \gls*{deberta} incorporates absolute positional information at the very end of the architecture instead of adding it to the token representations in the beginning as done with \gls*{bert}. Hence, the model is more accurately tuned to capture information on content and relative positions as only a small amount of parameters attends to the absolute positioning of tokens.

\textbf{\Gls*{luke}.} \gls*{luke} \cite{yamada_luke_2020} extends the masked \gls{lm} of BERT with a novel training task that accounts for entity-level tokens. This is achieved by explicitly masking entities in addition to words and giving them a distinct token embedding. Furthermore, the attention mechanism is adapted to explicitly model the relations of words to entities and vice versa. \Gls*{luke} also accounts for architectural optimization since \gls*{bert} by using the improved \gls*{roberta} as its baseline.

\textbf{Switch Transformer.} The Switch Transformer \cite{JMLR:v23:21-0998} represents a sparsely-activated model that maximizes parameter count efficiently to increase model performance, while simultaneously reducing training time. More accurately, the model consists of a large number of expert models with their own parameters. In each layer the single best expert is determined for each token in the input by a routing mechanism and the resulting representations are linearly combined and passed on to the next layer. Since the size of all weight matrices is determined beforehand but the decisions of the routing mechanism are dynamic, an auxiliary loss is added to the model to encourage the router to evenly distribute the layer inputs across all available experts. Otherwise, too many tokens may be assigned to an expert, which causes those tokens to be dropped, that is not be used in the current layer.

\textbf{\Gls*{kepler}.} \gls*{kepler} \cite{10.1162/tacl_a_00360} captures factual as well as linguistic knowledge in its embeddings by training on knowledge graphs and arbitrary text respectively. Concretely, the training of \gls*{kepler} is jointly governed by two loss functions expressed in a knowledge embedding task with negative sampling and the masked language modeling task of \gls*{bert}. The knowledge embedding task uses entity descriptions and fixed relation embeddings extracted from knowledge graphs and subsequently maximizes the similarity of a given entity description with any neighboring entity description, while minimizing the similarity of not connected entities. At that, the projection of the entity descriptions is done with the same encoder that is used during masked language modeling. Hence, \gls*{kepler} can build on top of existing Transformer architectures. Precisely, \gls*{kepler} initializes with a \gls*{roberta} checkpoint.

\textbf{\Gls*{canine}.} \citet{10.1162/tacl_a_00448} identify explicit tokenization as a rudiment of the beginnings of artificial-neural-network-based \gls*{tr}. They present \gls*{canine}, an encoder-only Transformer that instead learns tokenization by operating directly on byte strings. 
In particular, Unicode character codepoints are converted into numerical representations by concatenating the results of multiple hashing functions. These representations are then fed to a block-wise local attention layer in order for it to learn a composition function to greater linguistic units, for example subwords.
Next, strided convolution is applied to the representation vectors as a downsampling method to compensate for the greater sequence length of Unicode character codepoints in comparison to textual units, for example characters. The resulting representations are then fed through a deep stack of Transformer encoders. For downstream tasks that require the prediction of text sequences, an upsampling method is provided. The downsampled deep representations of the model are padded to the input sequence length by replicating them $n$ times, $n$ being the downsampling rate, and concatenating them with the representations of the local attention layer.

\subsection{Genealogy}
In order to trace the history of the presented \gls*{tr} methods we place them in a genealogy that comprises temporal information along one axis and identifies different evolutionary branches on the other axis. The latter describes four paradigms that can be traced throughout the history of \gls*{tr}: \textit{size}, \textit{context}, \textit{efficiency}, and \textit{multi-tasking}. Hence, any \gls*{tr} method can be assigned to at least one of these evolutionary branches. 

\textbf{Size.} \gls*{tr} methods that fall into the \textit{size} branch value more data and larger models over highly curated corpora and custom-built training tasks. These models consistently achieve state-of-the-art results on \gls*{nlp} tasks but require extensive computational resources and distributed architectures, making them inaccessible for most researchers.

\textbf{Context.} The representatives of the \textit{context} branch aim at increasing the distance of textual dependencies that can be modeled. This is achieved either by allowing for longer text sequences to be processed by a model or by adapting the memory mechanism of a model, for example storing more network activations. Context information is especially valuable if either the input or the output of a model is expected to be long in a given \gls*{nlp} task, for example for summarization.

\textbf{Efficiency.} The \textit{efficiency} branch tries to achieve high performance on a small scale. It has gained traction due to the trend of continuously creating larger, more potent models, which, however, remain inaccessible for most researchers due to immense resource costs. Common approaches are the optimization of the operations of a \gls*{tr} method or the transferal of the abilities of larger models to versions with a smaller footprint.

\textbf{Multi-tasking.} Finally, the \textit{multi-tasking} branch aims at explicitly capturing many facets of language by simultaneously optimizing a model on various carefully crafted training tasks. Hence, finding an effective combination of training tasks is the principal objective of this branch. For instance, it could include semantic, syntactic, and orthographic tasks.

Figure \ref{fig:genealogy} provides a schematic overview of the relations of the different \gls*{tr} methods.



\Figure[!t]()[width=0.70\textwidth]{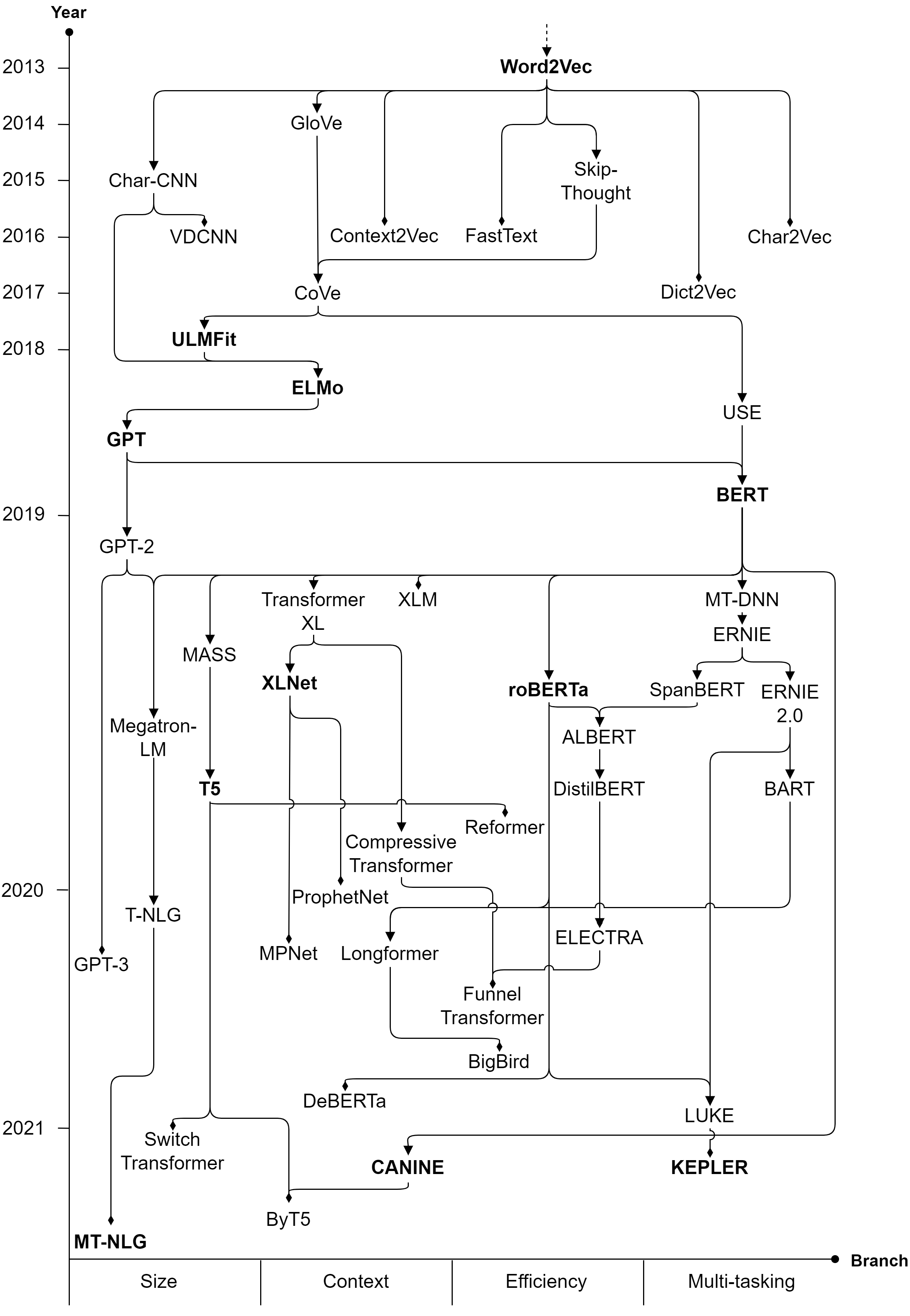}
   {Genealogy of recent \acrshort*{tr} methods.\label{fig:genealogy}}

An analysis of the temporal axis of the genealogy indicates that the evolution of \gls*{tr} can be split into two phases. The \textit{first phase} ranges from 2013 to 2017 and the \textit{second phase} ranges from 2019 until today. 2018 can be seen as the transition between the two phases. At the beginning of each phase stands a novel approach that changed the landscape of \gls*{tr} and inspired the development of manifold \gls*{tr} methods. This is reflected in the high number of outgoing connections of such a \gls*{tr} method in the genealogy. Consequently, the first phase was initiated by the word2vec models and the second phase by \gls*{bert} and, in part, \gls*{gpt}. 
During the first phase, new \gls*{tr} methods were inspired by word2vec models, yet they scarcely built on top of each other and rather branched out separately to address the four paradigms size, context, efficiency, and multi-tasking. The high amount of leaf nodes in the genealogy illustrates that the first phase was characterized by radical changes concerning model architecture and other conceptional choices.
The second phase stands in stark contrast. While the \gls*{tr} models still branched out, the tendency was to refine and extend previous methods, in particular \gls*{bert} and \gls*{gpt}. This emphasizes that a robust architecture was found with the Transformer \cite{NIPS2017_3f5ee243}. 
An analysis of the transition period reveals two mostly disjoint \textit{streams} that clearly show the adoption of \gls*{lm} in favor of previous contextualization approaches. Moreover, a split can be identified between \gls*{tr} methods that employ autoregressive approaches (left) and those that use autoencoder approaches (right). Taking another look at the second period, however, the two split streams are in the process of reuniting due to an increasing number of new sequence-to-sequence architectures, which combine aspects of autoencoding and autoregression. 

Finally, an illustration of the central problems in the history of \gls*{tr} is helpful to comprehend the motivation that drove the evolution of \gls*{tr} methods. In our genealogy, we highlight methods in bold that constitute the solution to such a problem and subsequently advanced the performance of \gls*{nlp} in general and \gls*{tr} in particular. It is worthwhile to mention that the transition period played a major role in shaping the evolution of \gls*{tr} as indicated by several consecutive bold methods.
The first hurdle of modern \gls*{tr} was a severe computational bottleneck. Pioneering \gls*{ffnn} could not efficiently model the highly variant feature space of natural language and had to either reduce computation time or training data, both leading to diminishing \gls*{tr} capabilities. The hurdle was taken with the \textbf{word2vec} models that provided a parameter efficient architecture to make the calculation of distributional embeddings feasible for large corpora.
However, the embeddings were stored in static lookup-tables. In effect, homonyms were united in a single embedding and thus made indistinguishable. The setup thus inherently discarded polysemy. In addition, the resulting embedding inevitably modeled a bias that was likely extreme and intransparent.
\textbf{\gls*{elmo}} received a lot of attention for solving the problem of polysemy. The approach created token embeddings on a per sample basis as a function of the entire input sequence. This allowed for the distinction of homonyms according to context information. Instead of inputting static lookup-tables, \gls*{elmo} provided a dynamic embedding model that could be prepended to a task-specific architecture.
However, a fundamental design principle for the training of artificial neural networks was still violated. The models could not directly map raw data to an inference, which gave space to errors and diminished performance. \textbf{\gls*{ulmfit}} proposed an end-to-end approach with great success and restructured the course of the \gls*{nlp} domain in its wake.
Recent years document that the Transformer has established itself as the best performing end-to-end architecture for \gls*{nlp}. With that the central problems in the \gls*{nlp} domain have shifted from changing the layout and combination of different model architectures to exploiting the capacities of the now ubiquitous Transformer. At the beginning of that process stand \textbf{\gls*{bert}} and \textbf{\gls*{gpt}} which achieved high language understanding and generation capabilities respectively. Subsequently, the creators of \textbf{XLNet} aspired to combine the generative power of \gls*{gpt} and discriminative power of \gls*{bert} into one model. \textbf{\gls*{roberta}} focused on refining \gls*{bert} to train more efficiently and effectively and managed to lay a strong foundation that has been used for the creation of new models up to more than a year later. \textbf{\gls*{kepler}} focused on explicitly integrating common sense and world knowledge to draw \gls*{nlp} methods near to the pragmatic layer. Meanwhile, \textbf{\gls*{t5}} was designed to be task-agnostic, that is  applicable to any \gls*{nlp} problem, as a one-for-all solution. \textbf{\gls*{canine}} extended the end-to-end principle to the model vocabulary by operating on byte strings instead of predefined token vocabularies. Lastly, \textbf{\gls*{mtnlg}} pushed computational boundaries with an immense amount of model parameters driven by the goal of creating an artificial intelligence that captures and thus understands all aspects of language.

\section{Discussion of Taxonomy of Text Representation Methods}

The compilation and analysis of \gls*{tr} methods conducted above enabled us to create a taxonomy of \gls*{tr} methods. It comprises the five dimensions \textit{architecture}, \textit{vocabulary}, \textit{representation}, \textit{domain-dependency}, and \textit{training strategy} along which a classification of a given \gls*{tr} method can be made (see Table \ref{tab:taxonomy}). In the following, we describe its dimensions and discuss the changes through time in each dimension to point out the current state-of-the-art. Note that the dimensions are sometimes fuzzy, for example \gls*{elmo} combines two architectures, the \gls*{cnn} and the \gls*{lstm}. A classification of all \gls*{tr} methods can be found in Table \ref{tab:classification}.


\begin{table*}[ht]
    
\begin{tabular}{|c|cccccccccccc}

\hline

\rowcolor{darkgray} \textcolor{white}{Dimension} & \multicolumn{12}{c|}{\textcolor{white}{Expression}}   
\\ \Xhline{1.3pt}

Architecture & \multicolumn{3}{c|}{ \acrshort{ffnn} } & \multicolumn{3}{c|}{ \acrshort{cnn} } & \multicolumn{3}{c|}{ \acrshort{rnn} } & \multicolumn{3}{c|}{\cellcolor{lightgray}Transformer}    
\\ \hline

Vocabulary & \multicolumn{3}{c|}{Byte}                & \multicolumn{3}{c|}{Character}                  & \multicolumn{3}{c|}{\cellcolor{lightgray} Subword}    & \multicolumn{3}{c|}{Word}  
\\ \hline

Representation &  \multicolumn{3}{c|}{Byte}                  & \multicolumn{2}{c|}{Character}    
& \multicolumn{2}{c|}{\cellcolor{lightgray}Subword} 
& \multicolumn{2}{c|}{Word}    
& \multicolumn{3}{c|}{Sequence}    
\\ \hline

Domain-dependency & \multicolumn{4}{c|}{\quad\quad Supervised\quad\quad\quad }                & \multicolumn{4}{c|}{\cellcolor{lightgray} Semi-supervised}                  & \multicolumn{4}{c|}{\quad Unsupervised \quad}    
\\ \hline

Training Strategy & \multicolumn{3}{c|}{\makecell{Context \\ Compression}}                & \multicolumn{3}{c|}{\makecell{Autoregressive \\ \acrshort{lm}}}                  & \multicolumn{3}{c|}{\cellcolor{lightgray}\colorbox{lightgray}{\makecell{Autoencoder \\ \acrshort{lm}}}}    & \multicolumn{3}{c|}{\gls{seq2seq}}  
\\ \hline

\end{tabular}

\caption{Taxonomy of \acrshort{tr} methods highlighting the state-of-the-art.}
\label{tab:taxonomy}
\end{table*}



\begin{table*}
  \rowcolors{4}{lightgray}{white}
  \begin{tabular}{|r?{1.3pt}*{4}{c}|*{4}{c}|*{5}{c}|*{3}{c}|*{4}{c}|}
      \multicolumn{1}{c?{1.3pt}}{} & \multicolumn{4}{c}{\cellcolor{darkgray} \textcolor{white}{Architecture}} & \multicolumn{4}{c}{\cellcolor{darkgray} \textcolor{white}{Vocabulary}} & \multicolumn{5}{c}{\cellcolor{darkgray}\colorbox{darkgray}{\textcolor{white}{\makecell{Representation}}}} & 
      \multicolumn{3}{c}{\cellcolor{darkgray}\colorbox{darkgray}{\textcolor{white}{\makecell{Domain- \\ dependency}}}} & 
      \multicolumn{4}{c|}{\cellcolor{darkgray}\colorbox{darkgray}{\textcolor{white}{\makecell{Training \\ strategy}}}} \\
  
    \cline{2-21}
    
    \multicolumn{1}{P{90}{3.0cm}?{1.3pt}}{} & \multicolumn{1}{P{90}{2.2cm}}{\acrshort{ffnn}} &
    \multicolumn{1}{P{90}{3.0cm}}{\acrshort{cnn}} &
    \multicolumn{1}{P{90}{3.0cm}}{\acrshort{rnn}} &
    \multicolumn{1}{P{90}{3.0cm}|}{Transformer} &
    \multicolumn{1}{P{90}{3.0cm}}{Byte} &
    \multicolumn{1}{P{90}{3.0cm}}{Character} &
    \multicolumn{1}{P{90}{3.0cm}}{Subword} &
    \multicolumn{1}{P{90}{3.0cm}|}{Word} &
    \multicolumn{1}{P{90}{3.0cm}}{Byte} &
    \multicolumn{1}{P{90}{3.0cm}}{Character} &
    \multicolumn{1}{P{90}{3.0cm}}{Subword} &
    \multicolumn{1}{P{90}{3.0cm}}{Word} &
    \multicolumn{1}{P{90}{3.0cm}|}{Sequence} &
    \multicolumn{1}{P{90}{3.0cm}}{Supervised} &
    \multicolumn{1}{P{90}{3.0cm}}{Semi-supervised} &
    \multicolumn{1}{P{90}{3.0cm}|}{Unsupervised} &
    \multicolumn{1}{P{90}{3.0cm}}{Context Compression} &
    \multicolumn{1}{P{90}{3.0cm}}{Autoregressive \acrshort{lm}} &
    \multicolumn{1}{P{90}{3.0cm}}{Autoencoder \acrshort{lm}} &
    \multicolumn{1}{P{90}{3.0cm}|}{\gls{seq2seq}} \\
    
    \Xhline{1.3pt}
    
    word2vec & X  &   &  &  &  &  &  & X && &  & X &  &  &  & X & X & & & \\
    \acrshort{glove} &  &   &  &  &  &  &  & X & && & X &  &  &  & X & X &  & &\\
    skip-thought & X  &   &  &  &  &  &  & X & && & & X &  & & X & X &  & &\\
    char-\acrshort{cnn} &   & X  &  &  &  & X & && &  &  & & X & X &  & & X &  & & \\
    FastText & X  &   &  &  &  & & X & & && X &  &  &  & & X & X &  & &\\
    char2vec &  &   & X &  &  & X &  && &  &  & X &  &  &  & X & X &  & &\\
    context2vec &   &   & X &  &  &  &  & X & && & X &  &  &  & X & X &  & &\\
    \acrshort{vdcnn} &   & X  &  &  &  & X & && &  &  &  & X & X &  &  & X &  &  &\\
    dict2vec & X  &   &  &  &  &  &  & X & && & X &  &  &  & X & X &  & &\\
    \acrshort{cove} &   &   & X &  &  &  & X & && & X  & & X &  &  & X &  &  & &\\
    \acrshort{ulmfit} &   &   & X &  &  &  &  & X & && & X &  &  & X &  &  & X & &\\
    \acrshort{use} & X  &   &  & X &  &  & & X & && & & X & X  &  & X & X &  & &\\
    \acrshort{elmo} &   & X  & X &  &  & X & && &  &  & X &  &  &  & X &  & X & &\\
    \acrshort{gpt}(1-3) & & &  & X &  &  & X & && & X &   &  &  & X &  &  & X & &\\
    \acrshort{bert}  &   &  &  & X &  &  & X & && & X &  & X &  & X &  &  &  & X &\\
    \acrshort{mtdnn} &   &  &  & X &  &  & X & && & X &  & X & X & X &  &   &  & X & \\
    \acrshort{xlm}  &   &   &  & X &  &  & X & && & X &  & X & X & X &  &  & X & X& \\
    TransformerXL   &   &  &  & X &  &  & X & && & X &  &  &  & X &  &  & X &  &\\
    \acrshort{ernie}(1-2)&   &   &  & X &  &  & X & && & X &  & X &  & X &  &  &  & X &\\
    MASS &   &   &  & X &  &  & X & && & X &  & &  & X &  &  &  & & X\\
    XLNet           &   &  &  & X &  &  & X & && & X &  & X &  & X &  &  & X & &\\
    \acrshort{roberta}&   &  &  & X &  &  & X & && & X &  & X &  & X &  &  &  & X &\\
    Span\acrshort{bert} &   &  &  & X &  &  & X & && & X &  & X &  & X &  &  &  & X &\\
    \acrshort{albert} &   &  &  & X &  &  & X & && & X &  & X &  & X &  &  &  & X &\\
    Megatron-LM &   &  &  & X &  &  & X & && & X &  & X &  & X &  &  & X & X &\\
    Distil\acrshort{bert} &   &  &  & X &  &  & X & && & X &  & X &  & X &  &  &  & X &\\
    \acrshort{t5} &   &   &  & X &  &  & X & && & X &  &  & X & X &  &  & & & X\\
    BART &   &  &  & X &  &  & X & && & X &  & X &  & X &  &  &  & & X\\
    Reformer      &   &   &  & X &  &  & X & && & X &  &  &  & X &  &  & & & X\\
    \colorbox{lightgray}{\makecell[r]{ Compressive \\ Transformer}}\hspace*{-0.7ex} &   &   &  & X &  &  & X & && & X &  &  &  & X &  &  & X & & \\
    ProphetNet &   &  &  & X &  &  & X & && & X &  & X &  & X &  &  & X & &\\
    T-NLG &   &  &  & X &  &  & X & && & X &  &  &  & X &  &  & X & &\\
    \acrshort{electra} &   &  &  & X &  &  & X & && & X &  & X &  & X &  &  &  & X &\\
    Longformer &   &  &  & X &  & X & X & & & X & X &  & X &  & X &  &  & X & X & X\\
    MPNet &   &  &  & X &  &  & X & && & X &  & X &  & X &  &  & X & X &\\
    \colorbox{lightgray}{\makecell[r]{ Funnel \\ Transformer}}\hspace*{-0.7ex} &   &  &  & X &  &  & X & && & X &  & X &  & X &  &  &  & X & X\\
    BigBird &   &  &  & X &  & & X & && & X &  & X & & X &  &  &  & X &\\
    DeBERTa &   &  &  & X &  &  & X & && & X &  & X &  & X &  &  & & X & \\
    LUKE &   &  &  & X &  &  & X & && & X &  & X &  & X &  &  &  & X &\\
    \colorbox{lightgray}{\makecell[r]{ Switch \\ Transformer}}\hspace*{-0.7ex} &   &  &  & X &  &  & X & && & X &  &  &  & X &  &  &  & X &\\
    KEPLER &   &  &  & X &  &  & X & && & X &  & X &  & X &  &  &  & X &\\
    CANINE &   &  &  & X & X &  & X & & X & & &  & X &  & X &  &  & X & X &\\
    ByT5 &   &  &  & X & X &  &  & & X &  &  &  &  & X & X &  &  &  & & X\\
    MT-NLG &   &  &  & X &  &  & X & && & X &  &  &  & X &  &  & X & &\\
    
    \hline
  \end{tabular}
  \caption{Classification of the identified \acrshort*{tr} methods according to our taxonomy.}
  \label{tab:classification}
\end{table*}


\textbf{Architecture.} The first dimension is the architecture of \gls*{tr} methods. The evolution in this dimension is mainly driven by the increase in computational capabilities and artificial neural networks. Early \gls*{tr} methods relied on \textit{\gls*{ffnn}} or therein inspired simple statistical calculations. The main limitation of these approaches were their inability to model sequence information efficiently. Due to their lack of expressiveness, these approaches have been replaced by deeper and more complex models, more precisely \textit{\gls*{cnn}} and \textit{\gls*{rnn}}. \gls*{cnn} stood out with their efficiency and ability to extract high-quality features from elementary parts of text, especially characters, while \gls*{rnn} capitalized on the sequential nature of text. \gls*{lstm} have had exceptional success by combining effective sequence modeling with stable network training. However, a step-change in \gls*{nlp} has been taking place with the introduction of the \textit{Transformer} architecture. It exploits present-day computational possibilities through its parallelizable structure, and outclasses other model architectures in terms of capturing long-range dependencies in text.

\textbf{Vocabulary.} The vocabulary dimension describes the tokenization granularity of a \gls*{tr} method. The body of literature distinguishes four granularities: \textit{byte}-level, \textit{character}-level, \textit{subword}-level, and \textit{word}-level tokens. On account of a straightforward implementation, the first approaches for \gls*{tr} used word-level tokens. However, the low granularity led to extensive vocabularies, resulting in generalization issues. A particular problem were \gls*{oov} and rare words. Character- and subword-level tokens subsequently mitigated the problem of large vocabularies and \gls*{oov} words but faced the challenge of learning meaningful compositions of tokens to larger linguistic units. Nonetheless, in terms of downstream task performance they surpassed word-level methods, implying the use of word morphology to better generalize language. More specifically, the morphological information could be accessed either implicitly through the combination of character-level features or explicitly through the incorporation of subword-level tokens into the vocabulary. However, character-level approaches struggled with underfitting natural language so that subword vocabularies have become the de facto standard. Importantly, this status quo might be about to change due to highly flexible byte-level approaches. They do away with explicit vocabulary creation and, in consequence, the need for text pre-processing, allowing models to operate on any language and to learn their own token boundaries in an end-to-end fashion at the cost of computational overhead. 

\textbf{Representation.} The representation dimension determines the units of text for which distributional representations are created by a \gls*{tr} method. A priori, the unit of representation of a given method cannot be of a finer granularity than its vocabulary. Contrarily, low-level token representations can be composed to represent larger linguistic units. Furthermore, the smallest linguistic unit distributional representations seem sensible for are subwords, given that morphemes are the smallest meaning bearing elements of language \cite{jurafsky_speech_2000}. Nonetheless, some approaches still choose a lower representation level. Overall, \gls*{tr} methods represent text on a \textit{byte}, \textit{character}, \textit{subword}, \textit{word}, and \textit{sequence} level. As with the vocabulary, \gls*{tr} started at the word-level because of the simple implementation. Afterwards, sequence representations were explored and performed better than word-level representations on \gls*{nlp} tasks that required such high-level representations, for example text classification. However, due to their aforementioned compositional flexibility, subword representations have become the de facto standard. Even so, most state-of-the-art \gls*{tr} methods additionally include sequence representations, because a simple composition of subwords to such a large textual unit would likely carry with it a degradation of the representation quality. That is, the meaning of a sequence can be larger than the combination of the meaning of the individual subwords \cite{goldberg_neural_2017}. This might further explain why character-level representations played only a marginal role in the evolution of \gls*{tr} methods. Note that while byte-level representation faces a similar problem, it benefits from the aforementioned advantages, that is  a vocabulary-free and thus more integrated end-to-end training. That being said, it remains to be seen whether byte-level can improve upon subword-level representations.

\textbf{Domain-dependency.} The fourth dimension of \gls*{tr} methods is the domain-dependency. \gls*{tr} can either be trained in a \textit{supervised}, \textit{semi-supervised}, or \textit{unsupervised} fashion. The first \gls*{tr} methods were unsupervised, which allowed the models to be trained on vast amounts of data. The resulting text representations embedded general aspects of language and could be used as the basis for task-specific architectures. Contrary to this development, supervised approaches were recognized to improve performance on \gls*{nlp} tasks by fitting text representations to the task domain. Yet, the evident drawback of supervised training was the limited in-domain data, and the approach was discarded as unsupervised corpora grew to effectively contain the entire textual internet. The necessity of training task-specific architectures from scratch when using unsupervised approaches on the one hand and the lack of training data in a supervised setting on the other hand, led to the adoption of semi-supervised \gls*{tr} methods. Semi-supervision combines the advantages of both previous approaches and hence established as the state-of-the-art in this dimension. A model is first trained on a huge corpus to capture the general aspects of language. Subsequently, the same model is fine-tuned on task-specific data. 

\textbf{Training strategy.} The last dimension is the training strategy. It refers to how the distributional hypothesis is implemented to create distributional representations. Four fundamental training strategies can be distinguished: \textit{context compression}, \textit{autoregressive \gls*{lm}}, \textit{autoencoder \gls*{lm}}, and \textit{\gls{seq2seq}}. The first approaches were \gls*{ffnn}-based strategies that aligned the vector representations of tokens that occurred together in fixed size windows around a position. 
These approaches were efficient but did not account for order information. Moreover, context-windows were typically smaller than the input sequence, limiting the context information available to these models.
Although later approaches mitigated both drawbacks by employing \gls*{rnn} and extending the context-window to the entire input sequence respectively, order information was still not used effectively. Specifically, merely compressing order and context information into embeddings according to a sequence was not sufficient.
Subsequently, autoregressive \gls*{lm} improved model performance by leveraging order information in explicitly learning the distribution of sequences in a corpus. However, these models introduced uni-directionality as a new limitation to the modeling of context information because only the preceding tokens to a target token were considered at any time-step. Many \gls*{tr} methods dealt with this problem by concatenating two opposite autoregressive \glspl*{lm} with good results, but autoencoder \glspl*{lm} made it possible to directly include bi-directional context information of an entire sequence for every position in a deep manner. Thereby, great strides were made in terms of downstream task performance on many \gls*{nlp} tasks, the caveat being inferior scalability and generative capabilities in comparison to autoregressive \glspl*{lm}. Hence, efforts were made to integrate bi-directional context into autoregressive \glspl*{lm}. Especially a hybrid learning strategy known as \gls{seq2seq} has gained traction. It combines aspects of autoencoding and autoregression for a flexible approach in terms of input manipulations, scalability, and downstream task applicability with competitive performance.

\section{Related Work}

In an attempt to survey \gls*{tr}, several authors have contributed to the field but did not provide a comprehensive compilation, composition, and systematization.

\citet{cambria_jumping_2014} use the concept of \textit{jumping curves} to illustrate the change of \gls*{nlp} from lexical to compositional semantics, that is  not analyzing the words but the concepts of a text. However, they do not cover specific \gls*{tr} methods.

\citet{9075398} break down \gls*{nlp} into natural language modeling, morphology, and semantics. The advances in each field are discussed in the context of deep learning and are framed with real world applications. Thus, the authors place a strong focus on task-specific models as well as architectures and digress from presenting the underlying \gls*{tr} methods. Furthermore, due to the publishing date of the paper, current advances are not covered.

\citet{ferrone_symbolic_2020} review the different modalities of \gls*{tr}, for example symbolic and distributed representation. They show a trade-off between expressiveness and interpretability. In particular, they examine the ability of distributed representations to exploit more aspects of texts, while being harder to interpret. Nevertheless, their research is limited to early \gls*{tr} methods.

\citet{10.1145/3495162} portray the evolution of \gls*{nlp} from shallow to deep models. They demonstrate this development with the example of the text classification task in \gls*{nlp}. While they mention a variety of different \gls*{tr} methods, the focus is rather placed on specific model architectures relating to downstream tasks.

\citet{zhou_progress_2020} structure the evolution of \gls*{nlp} along the perspectives modeling, learning, and reasoning. Similarly to \citet{10.1145/3495162}, they give a representative overview of \gls*{tr}, only briefly covering concrete methods. Moreover, reasoning is not related to \gls*{tr} specifically but rather to the broader \gls*{nlp} domain.

\citet{wang_survey_2020} identify three challenges in the history of \gls*{tr} methods: \gls*{oov} words, the understanding of context, and representations for different languages. They continue to show in detail the key \gls*{tr} methods and concepts that helped to overcome these challenges. However, improvements beyond the key approaches are not accounted for. As a result, we adopt a representative perspective, although more fine-grained with respect to \gls*{tr} than given by \citet{zhou_progress_2020} or \citet{10.1145/3495162}.

\citet{wang_static_2020} focus on illustrating the trend from static to dynamic \gls*{tr}. They extend the considerations from \citet{ferrone_symbolic_2020} by describing a selection of important \gls*{tr} methods in detail, including recent approaches. Moreover, they highlight the conceptual challenges the \gls*{tr} methods overcome to improve on previous approaches. Nonetheless, as in \cite{wang_survey_2020}, fine-grained developments in the field of \gls*{tr} are not covered.

\citet{peng_survey_2021} and \citet{duan_study_2020} elaborate on some of the most pervasive, recent \gls*{tr} methods. Yet, the brevity of the surveys is indicative of their incompleteness.

In contrast, \citet{HAN2021225} provide an extensive review of pre-trained models for \gls*{tr}. More precisely, they group the models according to the underlying problem that motivated their conception. The identified groups are further used to visualize the interrelations between and motivations behind the models in a family tree. 
But in spite of the level of detail throughout the paper, the authors shed little light on approaches preceding \gls*{gpt} and \gls*{bert}, do not explain how the presented models work in detail, and show only rudimentary interrelations between the models. 

\section{Conclusion}

\gls*{tr} has been evolving at an unprecedented rate. The result is a plethora of new and improved \gls*{tr} methods leading to a diffuse overall picture of the research field, which is confusing for novices in the domain of \gls*{nlp} and at least challenging for experts to keep track of. In this work, we made the effort to compile and systematize these \gls*{tr} methods and thus shed light on the recent evolution of \gls*{tr}.

In order to trace the genealogy of \gls*{tr}, many perspectives on the lineage of \gls*{tr} methods are possible. A comprehensive view can be achieved through their combination. In particular, we described the evolution through the development of conceptual choices, chronological relations, and design paradigms. In addition, the motivations behind pivotal \gls*{tr} methods supplement more detailed information.
The end of the genealogy displays the current state-of-the-art. However, it cannot be defined as singular methods but rather as a combination of conceptual choices. More concretely, Transformer-based models that train a semi-supervised \gls*{lm} on subwords to form subword representations.

Looking forward it can be suspected that the pace of radical changes in the \gls*{tr} domain, which has been sustained over recent years, is decelerating. The Transformer architecture seems to have stopped conceptual revolutions in favor of gradual increments. Nevertheless, new approaches are constantly being proposed introducing changes that need to be evaluated. 

Current developments suggest that progress will be divided into two fields. On the one hand, there is an increased interest in products based on artificial intelligence, which drives the development of task-specific architectures, models, and methods. On the other hand, resource-rich players, for example Google and Nvidia, continue to search for breakthroughs on the research side of \gls*{tr}. In particular, a complete understanding of natural language, that is  on the lexical, syntactic, semantic and, crucially, pragmatic layer.

Unfortunately, the necessity for an enormous amount of computational resources in the latter case is closing off the research field for many. In light of this, focus may as well shift to other open issues, for example explaining distributional representations or dealing with structural bias.

\section*{Acknowledgement}
This research was partially supported by the Federal Ministry of Education and Research (BMBF) within the project "White-Box-AI" (grant number 01IS22080) and managed by the project management agency Deutsches Zentrum für Luft- und Raumfahrt e.~V. (DLR), DLR-Projektträger.


\bibliographystyle{IEEEtranN}
\bibliography{bibliography2}

\begin{IEEEbiography}[{\includegraphics[width=1in,height=1.25in,clip,keepaspectratio]{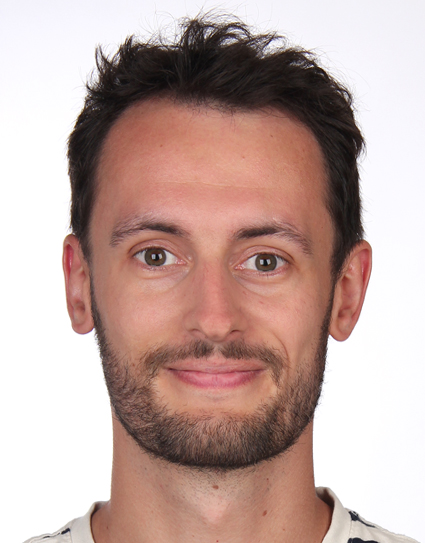}}]{Philipp Siebers} completed his studies of Business Informatics at the Technische Universität Dresden, where he specialized in the field of Data Science. Having provided solutions for several computer vision projects with industry stakeholders, most recently, he focused on the field of natural language processing, which not only provided the framework for his thesis, but is currently being applied to power a start-up platform aimed at continued learning.
\end{IEEEbiography}

\begin{IEEEbiography}[{\includegraphics[width=1in,height=1.25in,clip,keepaspectratio]{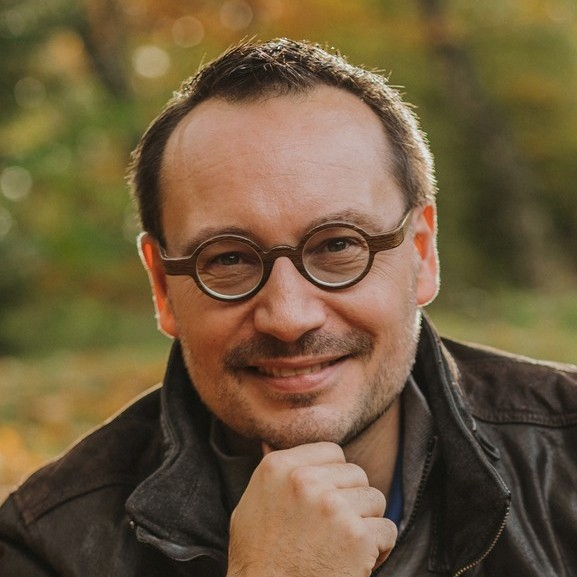}}]{Christian Janiesch} is Full Professor of Enterprise Computing at TU Dortmund University. Before, Christian worked full-time at various institutions including the Westfälische Wilhelms-Universität in Münster, the SAP Research Center Brisbane of SAP Australia Pty Ltd, the \mbox{Karlsruhe} Institute of Technology, and the Julius-Maximilians-Universität Würzburg. His research focuses on intelligent systems at the intersection of business process management and artificial intelligence with frequent applications in the Industrial Internet of Things. He is on the BPM Department Editorial Board for BISE journal, as well as the editorial boards of IJMR and JBA. He has authored over 150 scholarly publications. His work has appeared in journals such as the Journal of the Association for Information Systems, Communications of the Association for Information Systems, International Journal of Information Management, Information \& Management, Business \& Information Systems Engineering, Information Systems, Decision Support Systems, Future Generation Computer Systems as well as in various major international conferences including ICIS, ECIS, BPM, WI, and HICSS and has been registered as U.S. patents.
\end{IEEEbiography}

\begin{IEEEbiography}[{\includegraphics[width=1in,height=1.25in,clip,keepaspectratio]{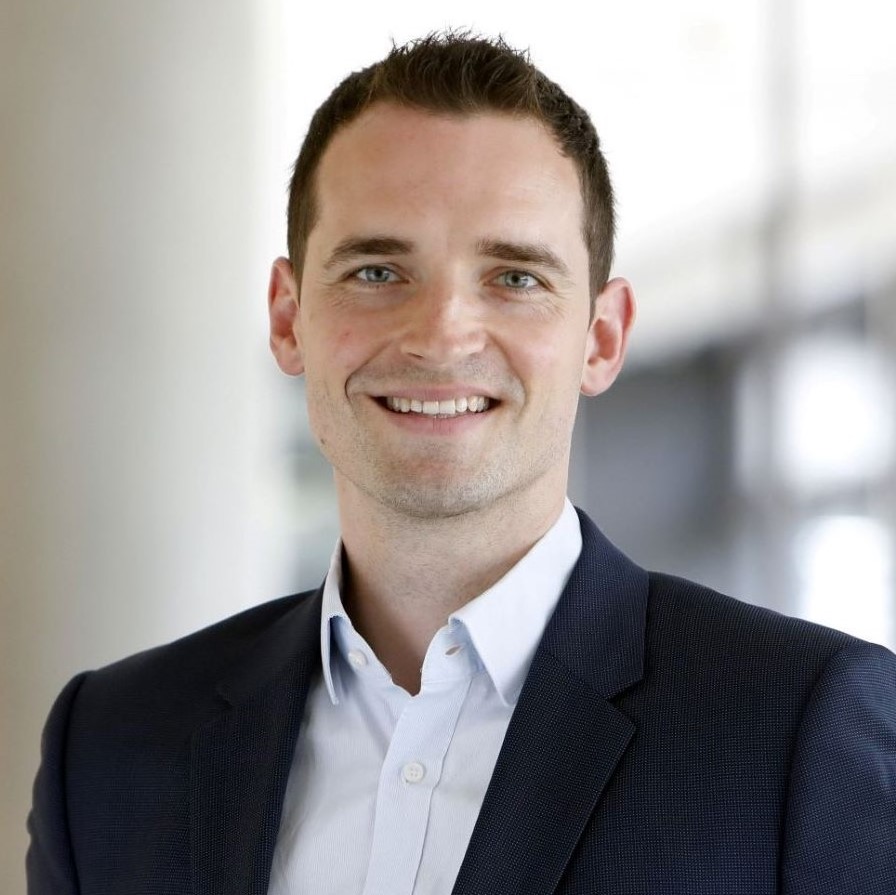}}]{Patrick Zschech} is Assistant Professor for Intelligent Information Systems at Friedrich-Alexander-Universität Erlangen-Nürnberg, Germany. Before that, he worked for the IT service provider Robotron Datenbank-Software GmbH as a project member and an instructor for data science qualification programs and received his PhD at the Technische Universität Dresden. The focus of his research is on business analytics, machine learning, and artificial intelligence with a particular interest in the design, analysis, and use of intelligent information systems. Patrick's results have been published in leading IS journals such as Decision Support Systems, Business \& Information Systems Engineering, Electronic Markets, and Journal of Business Analytics, and have been presented at international conferences such as ICIS, ECIS, and WI.

\end{IEEEbiography}

\EOD

\end{document}